\pdfoutput=1

\documentclass[11pt]{article}

\usepackage{aCL2023}

\usepackage{times}
\usepackage{latexsym}

\usepackage[T1]{fontenc}

\usepackage[utf8]{inputenc}

\usepackage{microtype}
\usepackage{inconsolata}

\usepackage{rotating}
\usepackage{blindtext}
\usepackage{amsmath}
\usepackage{amsfonts}
\usepackage{graphicx}
\usepackage{mathtools}
\usepackage{booktabs}
\usepackage{relsize}
\usepackage{float}
\usepackage{graphics}
\usepackage{multirow}
\usepackage{algorithm}
\usepackage{algpseudocode}
\usepackage[utf8]{inputenc}

\DeclareUnicodeCharacter{01B0}{\`u}
\DeclareUnicodeCharacter{01A1}{\`o}

\usepackage{threeparttable}
\usepackage{booktabs, caption, makecell}

\usepackage[title]{appendix}
\usepackage[cal=cm]{mathalfa}
\usepackage[colorinlistoftodos]{todonotes}
\usepackage{multirow}
\usepackage[export]{adjustbox}
\usepackage{subcaption}
\usepackage{pifont}
\newcommand{\mycomment}[1]{}                     
\newcommand{\cev}[1]{\reflectbox{\ensuremath{\vec{\reflectbox{\ensuremath{#1}}}}}}
\newcommand{\ignore}[1]{}

\newcommand{\cmark}{\ding{51}}%
\newcommand{\xmark}{\ding{55}}%
\usepackage{microtype}
\usepackage[subtle]{savetrees}
\title{Syntax-Aware Graph-to-Graph Transformer for Semantic Role Labelling}

\author{Alireza Mohammadshahi \\ 
        Idiap Research Institute \& EPFL 
        \AND 
        James Henderson\\
         Idiap Research Institute  \\
  \texttt{\{alireza.mohammadshahi, james.henderson\}@idiap.ch} \\ }

\begin{document}
\maketitle

\begin{abstract}

  
  Recent models have shown that incorporating syntactic knowledge into the semantic role labelling (SRL) task leads to a significant improvement. In this paper, we propose Syntax-aware Graph-to-Graph Transformer (SynG2G-Tr) model, which encodes the syntactic structure using a novel way to input graph relations as embeddings, directly into the self-attention mechanism of Transformer.  This approach adds a soft bias towards attention patterns that follow the syntactic structure but also allows the model to use this information to learn alternative patterns.  We evaluate our model on both span-based and dependency-based SRL datasets, and outperform previous alternative methods in both in-domain and out-of-domain settings, on CoNLL 2005 and CoNLL 2009 datasets.\footnote{The implementation is publicly available at \url{https://github.com/alirezamshi/SynG2GTr-SRL}.}

\end{abstract}

\section{Introduction}

The task of semantic role labelling (SRL) provides a shallow representation of the semantics in a sentence, and constructs event properties and relations among relevant words.
Traditionally, a syntactic structure was considered a prerequisite for SRL models~\cite{punyakanok-etal-2008-importance,gildea-palmer-2002-necessity}, but, newer models that leverage deep neural network architectures~\cite{cai-etal-2018-full,tan2017deep,he-etal-2017-deep,marcheggiani-etal-2017-simple} have outperformed syntax-aware architectures, without the need for explicit encoding of syntactic structure.
\ignore{
\begin{figure}
  \centering
  \includegraphics[width=\linewidth]{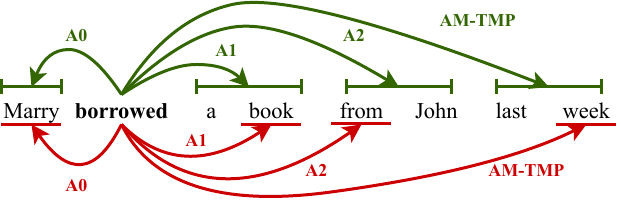}
  \caption{Example of SRL graphs. The upper structure is in the span-based style, and the lower one is in the dependency-based style.}
  \label{fig:examplesrl}
\end{figure}
}
However, recent studies~\cite{zhou-etal-2020-parsing,strubell-etal-2018-linguistically,he-etal-2017-deep,marcheggiani-titov-2017-encoding} have proposed that deep neural network models could benefit from using syntactic information, rather than disregarding it. These studies suggest that incorporating syntax into the model can improve SRL prediction by jointly learning both syntactic and semantic structures~\cite{zhou-etal-2020-parsing}, training a self-attention head in Transformer~\cite{NIPS2017_3f5ee243} to attend to each token's syntactic parent~\cite{strubell-etal-2018-linguistically}, or encoding the syntactic structure using graph convolutional networks~\cite{Fei_Li_Li_Ji_2021,marcheggiani-titov-2017-encoding}.\\
In this paper, we propose a novel method for encoding syntactic knowledge by introducing Syntax-aware Graph-to-Graph Transformer (SynG2G-Tr) architecture. The model conditions on the sentence's dependency structure and jointly predicts both span-based and dependency-based SRL structures. Inspired by \newcite{mohammadshahi2020recursive,mohammadshahi-henderson-2020-graph}, our model inputs graph relations as embeddings incorporated into the self-attention mechanism of Transformer~\cite{transformervaswani}. Different from the original Graph-to-Graph Transformer, our self-attention function models the interaction of the graph relations with both the query and key vectors of self-attention mechanism, instead of just the query. We also find that excluding the interaction of graph structure with the value vectors of self-attention does not harm the performance. Furthermore, compared to the previous work on Graph-to-Graph Transformers~\cite{mohammadshahi2020recursive,mohammadshahi-henderson-2020-graph}, our architecture uses different types of graphs as the input and output. 
We show empirically that our model outperforms previous comparable models. In the in-domain setting, SynG2G-Tr model achieves 88.93 (87.57)~F1 score on the CoNLL 2005 dataset, given the predicate~(end-to-end), and 91.23 (88.05)~F1 on the CoNLL 2009 dataset, given the predicate~(end-to-end). In the out-of-domain setting, our model reaches 83.21 (80.53)~F1 score on the CoNLL 2005 dataset, given predicate~(end-to-end), and 86.43~(81.93)~F1 scores on the CoNLL 2009 dataset, given predicate~(end-to-end).\\
Our contributions are:
\begin{itemize}
\item
    We propose SynG2G-Tr model for encoding the dependency parsing graph in the SRL task.
\item 
    We evaluate our model on CoNLL 2005 and CoNLL 2009 datasets and outperform previous comparable models in most cases of both in-domain and out-of-domain sets.
\end{itemize}

\begin{table}
\centering
  \begin{adjustbox}{width=0.7\linewidth}
  \begin{tabular}{lccc}
    \toprule
    Section & UAS & LAS & PoS \\
    \midrule
    Development & 96.72 & 94.83 & 96.81 \\
    Test & 96.85 & 95.24 & 97.41 \\
    \bottomrule
  \end{tabular}
  \end{adjustbox}
  \caption{Labelled and unlabelled attachment scores (LAS/UAS) and PoS accuracy. Sections 22\&23 of WSJ Penn Treebanks~\cite{marcus-etal-1993-building} are used as evaluation and test sets.\label{syn:acc}}
\end{table}
\section{Syntax-aware Graph-to-Graph Transformer}

\begin{figure}
  \centering
  \includegraphics[width=\linewidth]{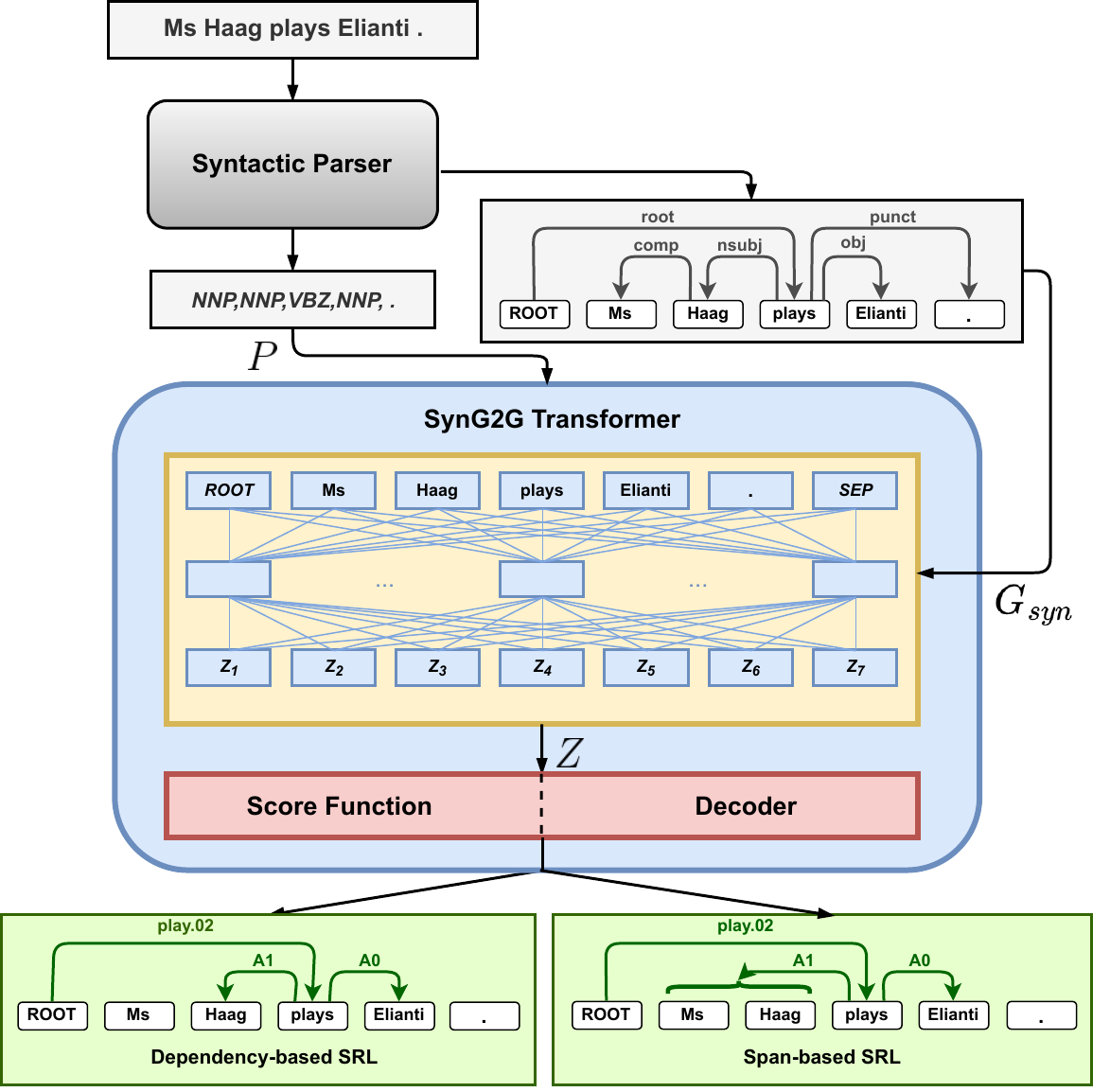}
  \caption{The architecture of SynG2G-Tr.}
  \label{fig:syng2g_model}
\end{figure}

The architecture of the SynG2G-Tr model is illustrated in Figure~\ref{fig:syng2g_model}. The input to the model is the tokenised text~($W=(w_1,w_2,...,w_N)$), which are the nodes of the input and output graphs, and $N$ is the length of tokenised input. The outputs are the dependency-based ($G_{dep}$) and span-based ($G_{span}$) SRL graphs. The SynG2G-Tr model can be formalised in terms of an encoder $E^{sg2g}$ and decoder $D^{sg2g}$:
\begin{equation}
    \begin{cases}
        Z = \operatorname{ E^{sg2g}}(W,P,G_{syn}) \\
        G_{span},G_{dep} = \operatorname{ D^{sg2g}}(Z)
    \end{cases}
\label{eq:syng2g-main}
\end{equation}
Initially, a syntactic parser predicts the dependency graph~($G_{syn}$), and Part-of-Speech (PoS) tags ($P=(p_1,p_2,...,p_N)$). Then the encoder of SynG2G-Tr~($E^{sg2g}$) encodes both sequences ($W, P$) and the dependency graph~($G_{syn}$) into contextualised representations of graph nodes~($Z$).
This representation ($Z$) is then used by the decoder~($D^{sg2g}$) to jointly predict SRL graphs. 
For the decoder, we follow the same unified scorer and decoder as defined in \newcite{zhou-etal-2020-parsing}. Further explanation of SRL scorer and decoding mechanism is provided in Appendix~\ref{app:score-decoder}.

The encoder employs an enhanced way of inputting graph relations into the self-attention mechanism of Transformer~\cite{transformervaswani}. Unlike the previously proposed version of Graph-to-Graph Transformer~\cite{mohammadshahi2020recursive}, we modify the self-attention mechanism to have a more comprehensive interaction between graph relations, queries and keys. We also find that excluding the interaction of graph relations with value vectors retains good performance.

\begin{figure*}
  \centering
  \includegraphics[width=\linewidth]{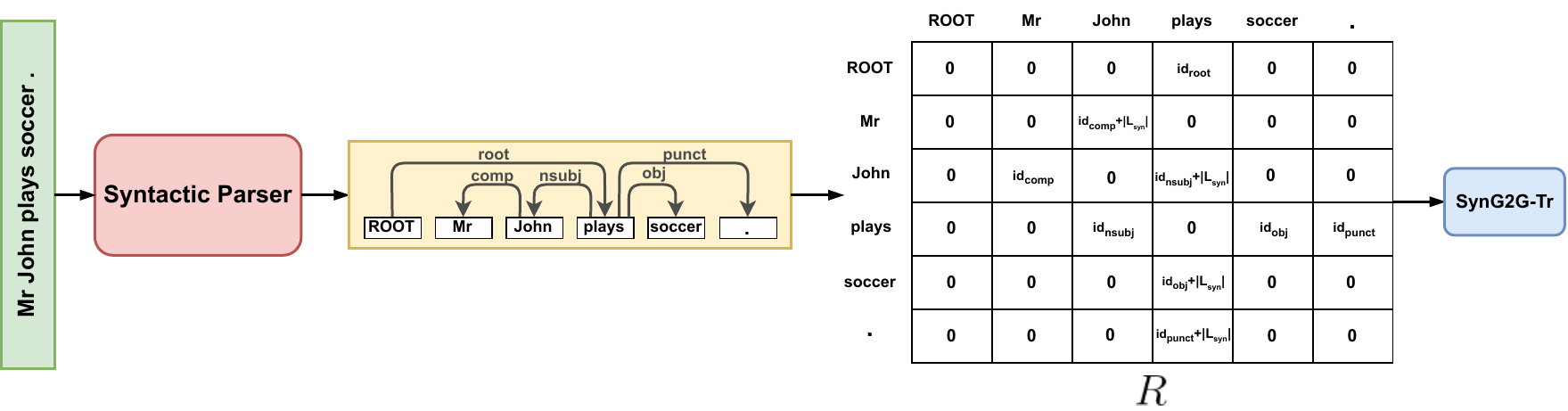}
  \caption{A sample computation of $R$ matrix for the sentence "Mr John plays soccer.".\label{fig:sample}}
  \label{fig:syng2g_sample}
\end{figure*}

Specifically, given the output of an intermediate embedding layer $X=(x_1,...,x_N)$, we define the attention mechanism of each head in each layer to take the dependency graph as input.  These attention scores~($\alpha_{ij}$) are calculated as a Softmax function over $e_{ij}$ values:
\vspace{-3ex}
\begin{align}
\label{eq:g2g-attn}
\\[-1ex]
\begin{split}
e_{ij} = \frac{1}{\sqrt{d}} \Big[ x_i\boldsymbol{W^Q}(x_j\boldsymbol{W^K})^T
+x_i\boldsymbol{W^Q}(r_{ij}\boldsymbol{W^R})^T \\[-1ex]
+r_{ij}\boldsymbol{W^R}(x_j\boldsymbol{W^K})^T \Big]
\end{split}
\nonumber
\end{align}
\ignore{
\begin{align}
\begin{split}
e_{ij} = \frac{1}{\sqrt{d}} \Big[ (x_i\boldsymbol{W^Q}+r_{ij}\boldsymbol{W^R})(x_j\boldsymbol{W^K}+r_{ij}\boldsymbol{W^R})^T \\
 -(r_{ij}\boldsymbol{W^R})(r_{ij}\boldsymbol{W^R})^T \Big]
\end{split}
\end{align}
}
where $\boldsymbol{W^Q},\boldsymbol{W^K} \in \mathbb{R}^{d_x \times d} $ are learned query and key matrices. $r_{ij} \in R$ is a one-hot vector specifying both the label and direction of the dependency relation between token $i$ and token $j$. $R$ is the matrix of graph relations, derived from the syntactic graph~($G_{syn}$). Figure~\ref{fig:sample} illustrates a sample computation of $R$ matrix, where $r_{ij} = id_{label}$ if $i \rightarrow j$, $id_{label}+|L_{syn}|$ if $j \leftarrow i$ , or {\sc none}~($|L_{syn}|$ is the size of syntactic label set). $\boldsymbol{W^R} \in \mathbb{R}^{(2|L_{syn}|+1)\times d}$ is a matrix of learned relation embeddings. $d$ is the attention head size, and $d_x$ is the hidden size. \\
The second and third terms in Equation~\ref{eq:g2g-attn} incorporate the graph information into the self-attention mechanism of Transformer with a soft bias, while the model can still learn other structures, using this encoded graph information. For better efficiency, we share the relation embeddings across multiple attention heads in each layer. Additionally, the computation complexity of both the second and third terms is $O(N)$, as we ignore the {\sc none} graph relation, and the syntactic dependency graph is a tree. The output of the attention function is the value embedding~($v_i$), which is calculated as:
\vspace{-1ex}
\begin{align}
\label{eq:g2g-value}
\begin{split}
v_i = \sum_j\alpha_{ij}(x_j\boldsymbol{W^V})
\\[-1ex]
\end{split}
\end{align}
which, in our model, does not use the graph, and $\boldsymbol{W^V} \in \mathbb{R}^{d_x \times d}$ is the learned value matrix.

\paragraph{Syntactic Parser.} The parser jointly predicts PoS tags and the dependency graph. We apply the parser defined in \newcite{zhou-etal-2020-parsing}, which uses a joint scorer and decoder for dependency and constituency graphs based on Head-driven Phrase Structure Grammar~\cite{zhou-zhao-2019-head}. This method has achieved state-of-the-art results in the dependency parsing task.

\section{Related Work}
\label{relatedwork}
Several approaches have been proposed to use syntax for the SRL task. \newcite{roth-lapata-2016-neural} embed dependency paths, while some researchers~\cite{Fei_Li_Li_Ji_2021,munir2021,marcheggiani-titov-2017-encoding} use graph convolutional networks to encode the syntactic structure. \newcite{strubell-etal-2018-linguistically} incorporates a dependency graph by training one attention head of Transformer to attend to syntactic parents for each token, in a multi-task setting. \newcite{he-etal-2019-syntax,he-etal-2018-syntax} use syntactic information to guide the argument pruning. \newcite{xia2019} exploit different alternatives e.g. tree-structured GRU and graph features of dependency tree to encode syntactic knowledge. \newcite{kasai-etal-2019-syntax} apply BiLSTM to tag the text with supertags extracted from dependency parses and feed them into SRL models. \newcite{xia-etal-2020-semantic} showed that incorporating heterogeneous syntactic knowledge results in significant improvement. Some other work focus on joint learning of both SRL and syntax~\cite{zhou-etal-2020-parsing,zhou-etal-2020-limit,cai-lapata-2019-semi,10.1162/tacl_a_00272}. Additionally, some approaches discarded the syntax, but achieve impressive results~\cite{shi2019simple,peters-etal-2018-deep,he-etal-2018-jointly,marcheggiani-etal-2017-simple,he-etal-2017-deep,tan2017deep,zhou-xu-2015-end}. \\
Our work is different from previous work since we encode the syntactic graph by directly inputting it as embeddings into the attention mechanism of Transformer, which provides a soft bias. Moreover, both sequences and syntactic graph can be encoded in one general model.
\section{Results and Discussion}

\paragraph{Experimental Setup.} Our models are evaluated on CoNLL 2005~\cite{carreras-marquez-2005-introduction} and CoNLL 2009~\cite{hajic-etal-2009-conll}.\footnote{Implementation details of datasets and SynG2G-Tr model are provided in Appendices~\ref{app:dataset} and \ref{app:hyper}.} For predicate disambiguation, we follow previous work~\cite{marcheggiani-titov-2017-encoding}, and use an off-the-shelf disambiguator from~\newcite{roth-lapata-2016-neural}. As in previous work, we evaluate in both \textit{end-to-end}, and \textit{given predicate} settings. 
For a more accurate comparison, we train SynG2G-Tr both with and without BERT initialisation~(SynG2G-Tr w/o BERT). The discrepancy between BERT tokenisation and the tokenisation used in the SRL corpora is handled as in \newcite{mohammadshahi-henderson-2020-graph}. 
\footnote{For inputting the dependency graph, the relation between two sub-words of different words is defined as the same dependency relation between their corresponding words in the sentence. This means that the relation ($i,j,l_{in}$) is repeated for each sub-word of word $x_i$, and word $x_j$. The same strategy is applied to predicted PoS tags.} For the syntactic parser, we use the same hyper-parameters as defined in \newcite{zhou-etal-2020-parsing}. The performance of the syntactic parser is shown in Table~\ref{syn:acc}. 

\begin{table}[t]
  \begin{adjustbox}{width=\linewidth}
  \begin{tabular}{lccccccccc}
    \toprule
    \multirow{2}{*}{Model} & \multirow{2}{*}{SA} & 
      \multicolumn{3}{c}{WSJ~(in-domain)} &&
      \multicolumn{3}{c}{Brown~(out-of-domain)}\\
      \cline{3-5} \cline{7-9}
     & & P & R & F1 && P & R & F1  \\
    \midrule
    \textbf{end-to-end}\\
    \newcite{he-etal-2017-deep} & \xmark & 85.0 & 84.3 & 84.6 && 74.9 & 72.4 & 73.6\\
    \newcite{he-etal-2018-jointly} & \xmark & 81.2 & 83.9 & 82.5 && 69.7 & 71.9 & 70.8 \\
    \newcite{strubell-etal-2018-linguistically} & \cmark & 85.53 & 84.45 & \underline{84.99} && 75.8 & 73.54 & \underline{74.66} \\
    \newcite{li2019dependency}  & \xmark & - & - & 83.0 && - & - & -\\
    \newcite{xia2019} & \cmark & 84.3 & 83.8 & 84.1 && 73.7 & 72.0 & 72.9 \\
    \newcite{xia-etal-2020-semantic} & \cmark &  83.05  & 84.49 & 84.49 &&  73.47 & 74.92 & 74.19 \\
    SynG2G-Tr (w/o BERT) & \cmark & 84.48 & 86.46 & \textbf{85.45} && 73.92 & 76.65 & \textbf{75.26} \\
    \hline
    \textit{+pre-training} \\
    \newcite{he-etal-2018-jointly} & \xmark & 84.8 & 87.2 & 86.0 && 73.9 & 78.4 & 76.1  \\
    \newcite{strubell-etal-2018-linguistically}$\dagger$ & \cmark & 87.13 & 86.67 & \underline{86.9} && 79.02 & 77.49 & \underline{78.25} \\
    \newcite{li2019dependency}  & \xmark & 85.2 & 87.5 & 86.3 && 74.7 & 78.1 & 76.4 \\
    SynG2G-Tr  & \cmark & 86.86 & 88.3 & \textbf{87.57} && 80.01 & 81.07 & \textbf{80.53} \\
    \midrule
    \textbf{given predicate} \\
    \newcite{tan2017deep} & \xmark & 84.5 & 85.2 & 84.8 && 73.5 & 74.6 & 74.1  \\
    \newcite{he-etal-2018-jointly} & \xmark & - & - & 83.9 && - & - & 73.7  \\
    \newcite{strubell-etal-2018-linguistically}$\dagger$ & \cmark & 86.02 & 86.05 & \underline{86.04} && 76.65 & 76.44 & \underline{76.54} \\
    \newcite{ouchi-etal-2018-span} & \xmark & 84.7 & 82.3 & 83.5 && 76.0 & 70.4 & 73.1  \\
    \newcite{xia-etal-2020-semantic} & \cmark &  85.12 & 85.0 & 85.06 && 76.3& 75.42 & 75.86 \\
    SynG2G-Tr (w/o BERT) & \cmark & 86.46 & 86.56 & \textbf{86.50} && 77.73 & 77.18 & \textbf{77.45} \\
    \hline
    \textit{+pre-training} \\
    \newcite{he-etal-2018-jointly} & \xmark & - & - & 87.4 && - & - & 80.4 \\
    \newcite{ouchi-etal-2018-span} & \xmark & 88.2 & 87.0 & 87.6 && 79.9 & 77.5 & 78.7  \\
    \newcite{li2019dependency}  & \xmark & 87.9 & 87.5 & 87.7 && 80.6 & 80.4 & 80.5\\
    \newcite{Jindal2020ImprovedSR} & \xmark & 87.70 & 88.15 & 87.93 && 81.52 & 81.36 & 81.44 \\
    \newcite{Zhang2021ComparingSE} & \xmark & 88.70 & 88.00 & 87.90 && 80.30 & 80.10 & 80.20 \\
    \newcite{Jia_Yan_Wu_Tu_2022} & \xmark & - & - & \underline{88.25} && - & - & \underline{81.90} \\
    SynG2G-Tr & \cmark & 89.11 & 88.74 & \textbf{88.93} && 83.93 & 82.50 & \textbf{83.21} \\
    \bottomrule
  \end{tabular}
  \end{adjustbox}
  \caption{Comparing our SynG2G-Tr with previous comparable models on CoNLL 2005 test sets. `SA' means a syntax-aware model. Scores being boldfaced means that they are significantly better than the second best model, specified by the underline marker.\label{srl-span-test}}
      \vspace{-1ex}
\end{table}

\paragraph{CoNLL 2005 Results.\footnote{Results are calculated with official evaluation scripts of CoNLL 2005~(\url{https://www.cs.upc.edu/~srlconll}).}} The results for span-based SRL are shown in Table~\ref{srl-span-test}.\footnote{For a fair comparison, we excluded \newcite{li-etal-2021-syntax,zhou-etal-2020-parsing}, as they use information from the constituency graph additional to the dependency tree. Also, to better understand the effect of syntactic information, we exclude \newcite{FERNANDEZGONZALEZ2023110127,zhou-etal-2022-fast,conia-navigli-2020-bridging}, as they exploited different scorer and training mechanism for SRL graphs. However, the best setting of SynG2G-Tr model still shows competitive or better results when compared to aforementioned excluded works.} Without BERT initialisation, our SynG2G-Tr model outperforms \newcite{strubell-etal-2018-linguistically}~(the second best model) in both \textit{end-to-end} and \textit{given-predicate} settings. This highlights the benefit of injecting the graph information into the self-attention mechanism using a soft bias, instead of hard-coding one attention head to attend to the syntactic parent of each token, as used in \newcite{strubell-etal-2018-linguistically}. The main reason for this improvement is that the model can still learn other attention patterns in combination with the graph information, which will be described later in this section. When adding BERT initialisation, our SynG2G-Tr model outperforms best previous work by 5.4\%/8.8\% F1 relative error reduction~(RER) on average in both in-domain and out-of-domain evaluation sets, which demonstrates the compatibility of the modified self-attention mechanism of SynG2G-Tr with BERT~\cite{devlin-etal-2019-bert} initialisation. 

\definecolor{darkblue}{rgb}{0, 0, 0.5}
\paragraph{CoNLL 2009 Results.\footnote{Scores are calculated with CoNLL 2009 shared task script~(\url{https://ufal.mff.cuni.cz/conll2009-st/}).}}

Table~\ref{srl-dep-test} illustrates the results of dependency-based SRL on the test set of CoNLL 2009 dataset. Without BERT initialisation, SynG2G-Tr significantly outperforms previous work in in-domain and out-of-domain settings. With BERT initialisation, our model significantly outperforms previous work in \textit{end-to-end} setting with 3.2\%/10.4\% F1 RER in both in-domain and out-of-domain evaluation sets, while having competitive performance in \textit{given-predicate} setting. For a better comparison with \newcite{Fei_Li_Li_Ji_2021}~(last setting of Table~\ref{srl-dep-test}), we also employ the gold dependency tree for training and use the predicted dependency graph at inference time. Our model significantly outperforms \newcite{Fei_Li_Li_Ji_2021}, especially on the out-of-domain dataset. This shows the benefit of encoding the dependency graph by modifying the self-attention mechanism of Transformer~\cite{transformervaswani} compared to using graph convolutional network, as in \newcite{Fei_Li_Li_Ji_2021}. 

\begin{table}[t]
  \begin{adjustbox}{width=\linewidth}
  \begin{tabular}{lccccccccc}
    \toprule
    \multirow{2}{*}{Model} & \multirow{2}{*}{SA} & 
      \multicolumn{3}{c}{WSJ~(in-domain)} &&
      \multicolumn{3}{c}{Brown~(out-of-domain)}\\
      \cline{3-5} \cline{7-9}
     & & P & R & F1 && P & R & F1  \\
    \midrule
    \textbf{end-to-end}\\
    \newcite{he-etal-2018-syntax} & \cmark & 83.9 & 82.7 & 83.3 && - & - & - \\
    \newcite{cai-etal-2018-full} & \xmark & 84.7 & 85.2 & 85.0 && - & - & \underline{72.5} \\
    \newcite{li2019dependency} & \xmark & - & - & \underline{85.1} && - & - & - \\
    SynG2G-Tr (w/o BERT) & \cmark & 84.10 & 87.07 & \textbf{85.59}  && 73.66 & 72.56 & \textbf{73.11} \\
    \hline
    \textit{+pre-training} \\
    \newcite{li2019dependency} & \xmark & 84.5 & 86.1 & \underline{85.3} && 74.6 & 73.8 & \underline{74.2} \\
    SynG2G-Tr & \cmark & 86.38 & 89.78 & \textbf{88.05} && 80.35 & 83.57 & \textbf{81.93} \\
    \midrule
    \textbf{given predicate} \\
    \newcite{marcheggiani-etal-2017-simple} & \xmark & 88.7 & 86.8 & 87.7 && 79.4 & 76.2 & 77.7 \\
    \textcolor{darkblue}{M\&T}\shortcite{marcheggiani-titov-2017-encoding} & \cmark & 89.1 & 86.8 & 88.0 && 78.5 & 75.9 & 77.2 \\
    \newcite{he-etal-2018-syntax} & \cmark & 89.7 & 89.3 & 89.5 && 81.9 & 76.9 & 79.3 \\
    \newcite{cai-etal-2018-full} & \xmark & 89.9 & 89.2 & \underline{89.6} && 79.8 & 78.3 & 79.0 \\
    \newcite{cai-lapata-2019-syntax} & \cmark & 90.5 & 88.6 & \underline{89.6} && 80.5 & 78.2 & \underline{79.4} \\
    \newcite{kasai-etal-2019-syntax} & \cmark & 89.0 & 88.2 & 88.6 && 78.0 & 77.2 & 77.6 \\\
    SynG2G-Tr (w/o BERT) & \cmark & 89.78 & 90.28 & \textbf{90.03} && 81.32 & 82.15 & \textbf{81.73} \\
    \hline
    \textit{+pre-training} \\
    \newcite{li2019dependency} & \xmark & 89.6 & 91.2 & 90.4 && 81.7 & 81.4 & 81.5 \\   
    \newcite{kasai-etal-2019-syntax} & \cmark & 90.3 & 90.0 & 90.2 && 81.0 & 80.5 & 80.8 \\
    \newcite{lyu-etal-2019-semantic} & \xmark & - & - & 90.99 && - & - & 82.18 \\
    \newcite{chen-etal-2019-capturing} & \xmark & 90.74 & 91.38 & \textbf{91.06} && 82.66 & 82.78 & 82.72 \\
    \newcite{he-etal-2019-syntax} & \cmark & 90.41 & 91.32 & 90.86 && 86.15 & 86.70 & \textbf{86.42} \\
    \newcite{cai-lapata-2019-semi} & \cmark &  91.1 & 90.4 & 90.7 && 82.1 & 81.3 & 81.6 \\
    \newcite{munir2021} & \cmark & 91.2 & 90.6 & 90.9 && 83.1 & 82.6 & 82.8 \\
    SynG2G-Tr & \cmark & 91.31 & 91.16 & \textbf{91.23} && 86.40 & 86.47 & \textbf{86.43} \\
    \hline
    \textbf{gold syntax} \\
    \newcite{Fei_Li_Li_Ji_2021} & \cmark & 92.5 & 92.5 & \underline{92.5} && 85.6 & 85.3 & \underline{85.4} \\
    SynG2G-Tr+Gold & \cmark & 92.71 & 93.37 & \textbf{93.03} && 88.27 & 88.31 & \textbf{88.29} \\
    \bottomrule
  \end{tabular}
  \end{adjustbox}
  \caption{Comparing our SynG2G-Tr with previous comparable models on CoNLL 2009 test sets. `SA' means a syntax-aware model. Scores being boldfaced means that they are significantly better than the second best model, specified by the underline marker.\label{srl-dep-test}}
  \vspace{-1ex}
\end{table}

\paragraph{Further Analysis.} We also analyse the self-attention matrix of SynG2G-Tr model for different heads and layers. Figure~\ref{fig:attn}~in Appendix~\ref{app:attn} demonstrates that the self-attention mechanism of SynG2G-Tr ignores the dependency graph information in the first few layers, and only uses the context-dependent information. However, as it progresses to upper layers, it begins to utilise the graph relation information, as shown in the attention matrix. This highlights the benefit of encoding the dependency graph with a soft bias as the model can still learn different structures in different layers, given this encoded graph information. Furthermore, in Appendix~\ref{app:ablation}, we show that removing the interaction of graph embeddings with key vectors results in a performance drop. Additionally, ignoring the interaction of graph relations with both key and query vectors~\footnote{This leads to a BERT-based syntax-agnostic model, similar to \newcite{shi2019simple}.} results in a significant drop as well. However, integrating the graph information into Equation~\ref{eq:g2g-value} as stated in \newcite{mohammadshahi2020recursive} does not improve the performance, and we remove it for better efficiency.

\section{Conclusion}

In this paper, we propose the Syntax-aware Graph-to-Graph Transformer architecture, which effectively incorporates syntactic information by inputting the syntactic dependency graph into the self-attention mechanism of Transformer. 
Our mechanism for inputting graph relation embeddings differs from the original Graph-to-Graph Transformer in that it models the complete interaction between the dependency relation, query vector and key vector. It also excludes the graph interaction with value vectors while maintaining good performance.
We have evaluated our model on CoNLL 2005 and CoNLL 2009 SRL datasets and outperformed previous comparable models. 
Future studies can apply our model to any NLP task which might benefit from conditioning on the syntactic structure or other graphs.
\section*{Limitations}
SynG2G-Tr encodes the syntactic dependency graph because the nodes of input and output graphs should be similar. Future work could include investigating the use of constituency graphs in the self-attention mechanism of Transformer~\cite{transformervaswani}, where the nodes of the input graph~(constituency graph) are different from those of the SRL output graph. In this paper, we initialise our model with the pre-trained BERT~\cite{devlin-etal-2019-bert} model. As future study, larger and better pre-trained language models will be used for the initialisation of SynG2G-Tr models, to achieve better performance. Additionally, future studies can easily extend our work to multilingual SRL benchmarks.

\section*{Acknowledgement}

We are grateful to the Swiss National Science Foundation~(SNSF), grant CRSII5-180320, for funding this work. We also thank members of the IDIAP NLU group for helpful discussions and suggestions. We are grateful to anonymous reviewers for their fruitful comments and corrections.


\bibliographystyle{acl_natbib}
\bibliography{emnlp2021}

\newpage
\renewcommand\thesection{\Alph{section}}
\renewcommand\thesubsection{\thesection.\Alph{subsection}}
\setcounter{section}{0}
\onecolumn
\begin{appendices}

\section{SRL Scorer and Decoder Details}
\label{app:score-decoder}

\paragraph{Scorer.} Inspired by \newcite{zhou-etal-2020-parsing}, we first define span representation ($s_{ij}$) as the difference between right and left end-points of the span:
\begin{align}
\label{eq:span-rep}
\begin{split}
s_{ij} = \vec{sr_j} - \cev{sl_i}
\end{split}
\end{align}
where $\vec{sr_j}$ is defined as $[\vec{z_{j+1}};\vec{z_j}]$, and $\cev{sl_i}$ is calculated as $[\cev{z_i};\cev{z_{i+1}}]$.  $\cev{z_i}$ is computed by dividing the output representation of Transformer~($z_i$) in half. \\
Argument~($a_{ij}$) and predicate~($v_k$) representations are defined as:
\begin{align}
\label{eq:arg-pred}
\begin{split}
&a_{ij} = \operatorname{ReLU}(\boldsymbol{W^1_{srl}}s_{ij}+ b^1_{srl}) \\
&v_k = z_k
\end{split}
\end{align}
where $\boldsymbol{W^1_{srl}}$ and $b^1_{srl}$ are learned parameters and $\operatorname{ReLU}(.)$ is the Rectified Linear Unit~\cite{10.5555/3104322.3104425} function.\\
We predict semantic roles as defined in \newcite{zhou-etal-2020-parsing}: 
\begin{align}
\label{eq:srl-score}
\begin{split}
\Phi_l(v,a) = \boldsymbol{W^3_{srl}}(\operatorname{LN}(\boldsymbol{W^2_{srl}}[a_{ij};v_k] + b^2_{srl}))+b^3_{srl}
\end{split}
\end{align}
where $\operatorname{LN}(.)$ is the layer normalisation~\cite{ba2016layer} function, and $\boldsymbol{W^2_{srl}}$, $\boldsymbol{W^3_{srl}}$, $b^2_{srl}$, and $b^3_{srl}$ are learned parameters.  The semantic role score for a specific label $l_{out}$ is defined as:
\begin{align}
\label{eq:srl-score2}
\begin{split}
\Phi_l(v,a,l_{out}) = [\Phi_l(v,a)]_{l_{out}} 
\end{split}
\end{align}
Since the number of predicate-argument pairs is $O(n^3)$, we apply the pruning method proposed in \newcite{li2019dependency,he-etal-2018-jointly} by defining separate scorers for argument and predicate candidates ($\Phi_a$ and $\Phi_v$), and pruning all but the top-ranked arguments and predicates based on their corresponding scores. 

\paragraph{Training.} The model is trained to optimise the probability $P(\hat{y}|W,P,G_{syn})$ of predicate-argument pairs, conditioned on input sequence~($W$), PoS tags~($P$), and predicated dependency graph~($G_{syn}$).  This objective can be factorised as:
\begin{align}
\label{eq:srl-loss}
\begin{split}
J(\theta) &= \sum_{y \in \Gamma} -log P_{\theta}(y|W,P,G_{syn})\\
&= \sum_{\langle v,a,l_{out}\rangle \in \Gamma} -log \frac{\operatorname{exp}(\Phi(v,a,l_{out}))}{\sum_{\hat{l} \in L_{srl}}\operatorname{exp}(\Phi(v,a,\hat{l}))}
\end{split}
\end{align}
where $\Phi(v,a,l_{out})$ is defined as $\Phi_v(v) + \Phi_a(a) + \Phi_l(v,a,l_{out})$, and $\theta$ is model parameters.  $\Gamma$ is the set of predicate-argument-relation tuples for all possible predicate-argument pairs and either the correct relation or {\sc none}.

\paragraph{Decoders.} Following \newcite{zhou-etal-2020-parsing}, we apply a single dynamic programming decoder according to the uniform score following the non-overlapping constraints~\cite{punyakanok-etal-2008-importance}.

\section{Implementation Details and Pre-processing Steps}
\label{app:dataset}

\paragraph{CoNLL 2005:} In this shared task~\cite{carreras-marquez-2005-introduction}~(under LDC license), the focus was on verbal predicates in English. The training data includes sections 2-21 of the Wall Street Journal (WSJ) dataset. Section 24 is considered as the development set, while section 23 is used as the in-domain test set. Three sections of the Brown corpus are used for the out-of-domain dataset. The dataset can be downloaded from \href{https://www.cs.upc.edu/~srlconll/}{here}, and pre-processing steps are provided in \href{https://github.com/strubell/preprocess-conll05}{here}. 

\paragraph{CoNLL 2009:} This shared task~\cite{hajic-etal-2009-conll}~(under LDC license) focused on dependency-based SRL and was created by merging PropBank and NomBank treebanks. We evaluate our models on the English dataset with the same split as the CoNLL 2005 dataset. The dataset and pre-processing steps can be found at \href{https://ufal.mff.cuni.cz/conll2009-st/}{here} and \href{https://github.com/strubell/conll09-preprocess}{here}. The number of sentences in train and evaluation sets is as follows:

\begin{table}[hbt!]
\centering
    \begin{minipage}{.5\linewidth}
        \begin{tabular}{|c|c|c|c|}
        \hline
      Train & Dev & Test-WSJ & Test-Brown \\
      \hline
      39'832 & 1'334 & 2'399 & 425 \\
      \hline
    \end{tabular}
    \end{minipage} 
    \caption{The number of sentences for each split of CoNLL 2005 and CoNLL 2009 datasets.}
\end{table}

\section{Hyper-parameters Setting}
\label{app:hyper}

We use \textit{bert-large-whole-word-masking}\footnote{\url{https://github.com/google-research/bert}. Apache License 2.0.}~(345M parameters) for the initialisation of encoder in SynG2G-Tr model. We use Adam optimiser~\cite{kingma2014adam} and apply separate optimisers for pre-trained parameters and randomly initialised ones. We use bucket batching, grouping sentences by their lengths to the same batch to speed up the model. Early stopping is used to mitigate over-fitting, as in previous work~\cite{mohammadshahi2022rquge,mohammadshahi-etal-2022-small,mohammadshahi-etal-2019-aligning}. In a pre-defined predicate setting, we use different dynamic programming decoders to find SRL graphs, since predicates are not necessarily the same in dependency-based and span-based SRL graphs. For choosing the best hyper-parameters, we use manual tuning to find the base learning rate and BERT learning rate. For other hyper-parameters, we follow previous work~\cite{zhou-etal-2020-parsing}. The base learning rate is selected from \{$1e-2$,$1e-3$,$1.5e-3$\}, and the BERT learning rate is chosen from \{$1e-5$,$1.5e-5$,$2e-5$\}. So, we train our models with 9 different learning rates to find the best performing model based on the summation of F1 scores of span-based and dependency-based SRL graphs. We use NVIDIA GeForce GTX 1080 Ti for training and evaluating our models.~\footnote{The training time of SynG2G-Tr model is 0h20m40s, and the evaluation time is 0h02m24s.} For the dependency parser, we apply the same hyper-parameters as \newcite{zhou-etal-2020-parsing}. We use the base learning rate of $2e-3$, and the BERT learning rate of $1.5e-5$. Here is the list of hyper-parameters for the SynG2G-Tr model:

\begin{table}[hbt!]
    \begin{minipage}{.5\linewidth}
      \centering
        \begin{tabular}{c|c}
      Component & Specification \\
      \hline
      \textbf{Optimiser} & Adam \\
      Base Learning rate & 1.5e-3 \\
      BERT Learning rate & 1e-5 \\
      Adam Betas~($b_1$,$b_2$) & (0.9,0.999) \\
      Adam Epsilon & 1e-5 \\
      Weight Decay & 0.01 \\
      Max-Grad-Norm & 1 \\
      Warm-up & 0.001 \\
      \hline
      \textbf{Self-Attention} \\
      No. Layers & 24 \\
      No. Heads & 16 \\
      Embedding size & 1024 \\
      Max Position Embedding & 512 \\
      \hline
        \end{tabular}
        \end{minipage}
    \begin{minipage}{.5\linewidth}
      \centering
    \begin{tabular}{c|c}
      Component & Specification \\
      \hline
      \textbf{Feed-Forward layers~(SRL)} \\
      Span Hidden size & 512 \\
      Label Hidden size & 250 \\
      \hline
      \textbf{Feed-Forward layers~(PoS)} \\
      Hidden size & 250 \\
      \hline
      \textbf{Pruning~(SRL)} \\
      $\lambda_{verb}$ & 0.6 \\
      $\lambda_{span}$ & 0.6 \\
      Max No. Span & 300 \\
      Max No. Verb & 30 \\
      \hline
      Epoch & 100 \\
      \hline
    \end{tabular}
    \end{minipage} 
    \caption{\label{suptab:wsjhyper} Hyper-parameters for training SynG2G-Tr.}
\end{table}

\section{Attention Visualisation}
\label{app:attn}

\begin{figure*}[htb!]
        \centering
        \begin{minipage}{0.6\linewidth}
        \begin{subfigure}[b]{0.475\textwidth}
            \centering
            \includegraphics[width=\textwidth]{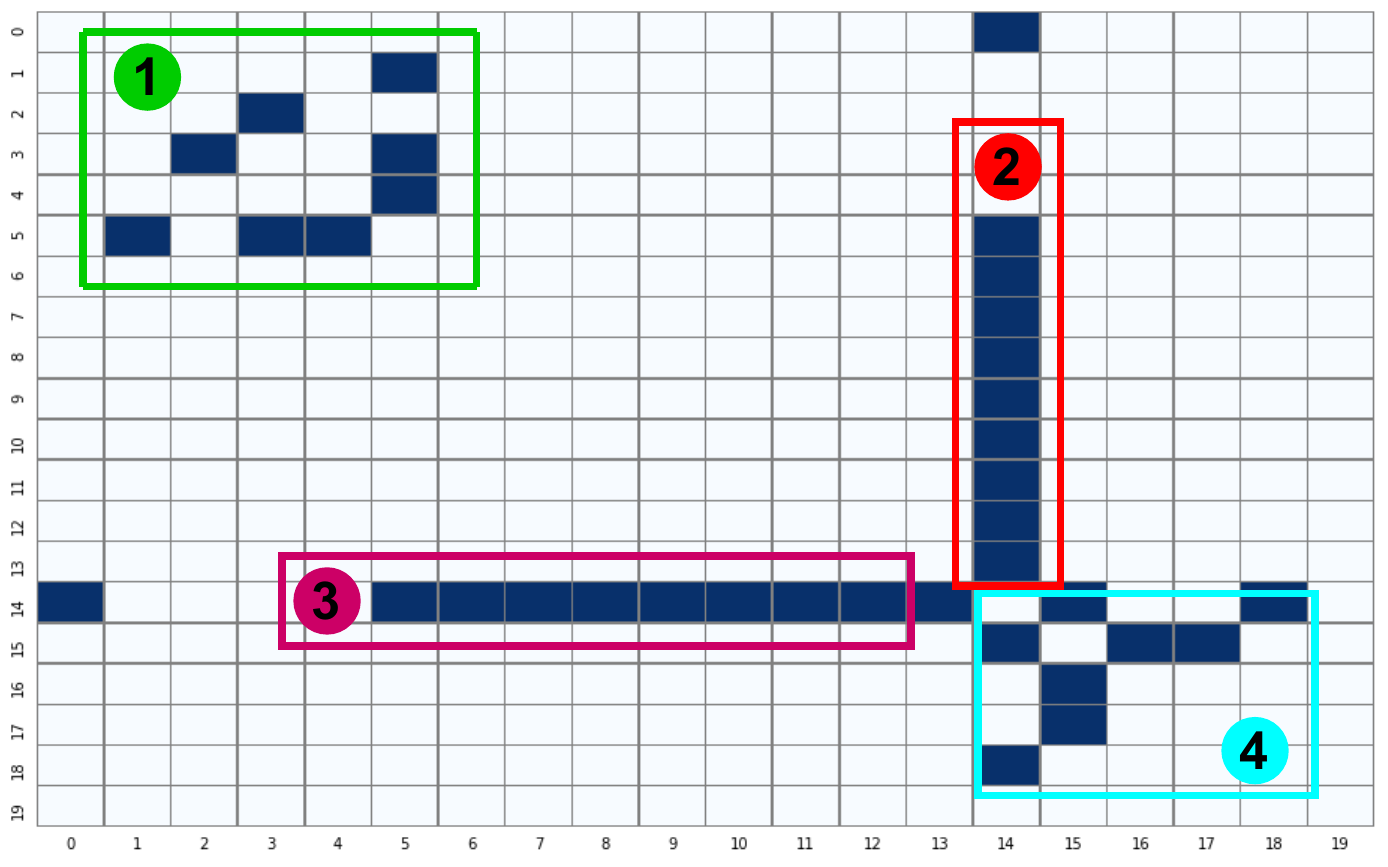}
            \caption[Graph Relation Matrix]%
            {{\small Graph Relation Matrix}}    
            \label{fig:attn1}
        \end{subfigure}
        \hfill
        \begin{subfigure}[b]{0.475\textwidth}  
            \centering 
            \includegraphics[width=\textwidth]{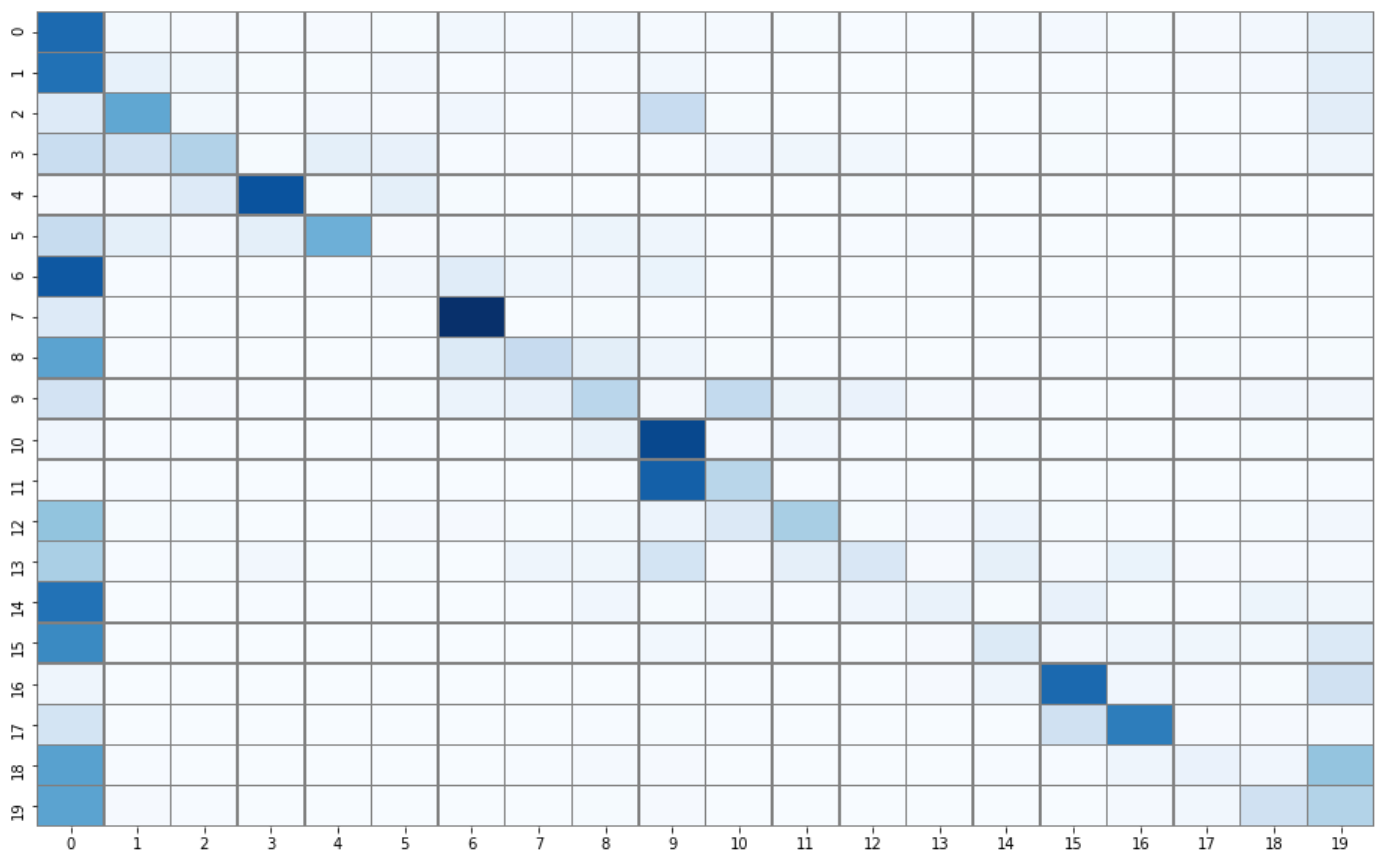}
            \caption[First Layer]%
            {{\small First Layer}}    
            \label{fig:attn2}
        \end{subfigure}
        \vspace{0.1cm}
        \begin{subfigure}[b]{0.475\textwidth}   
            \centering 
            \includegraphics[width=\textwidth]{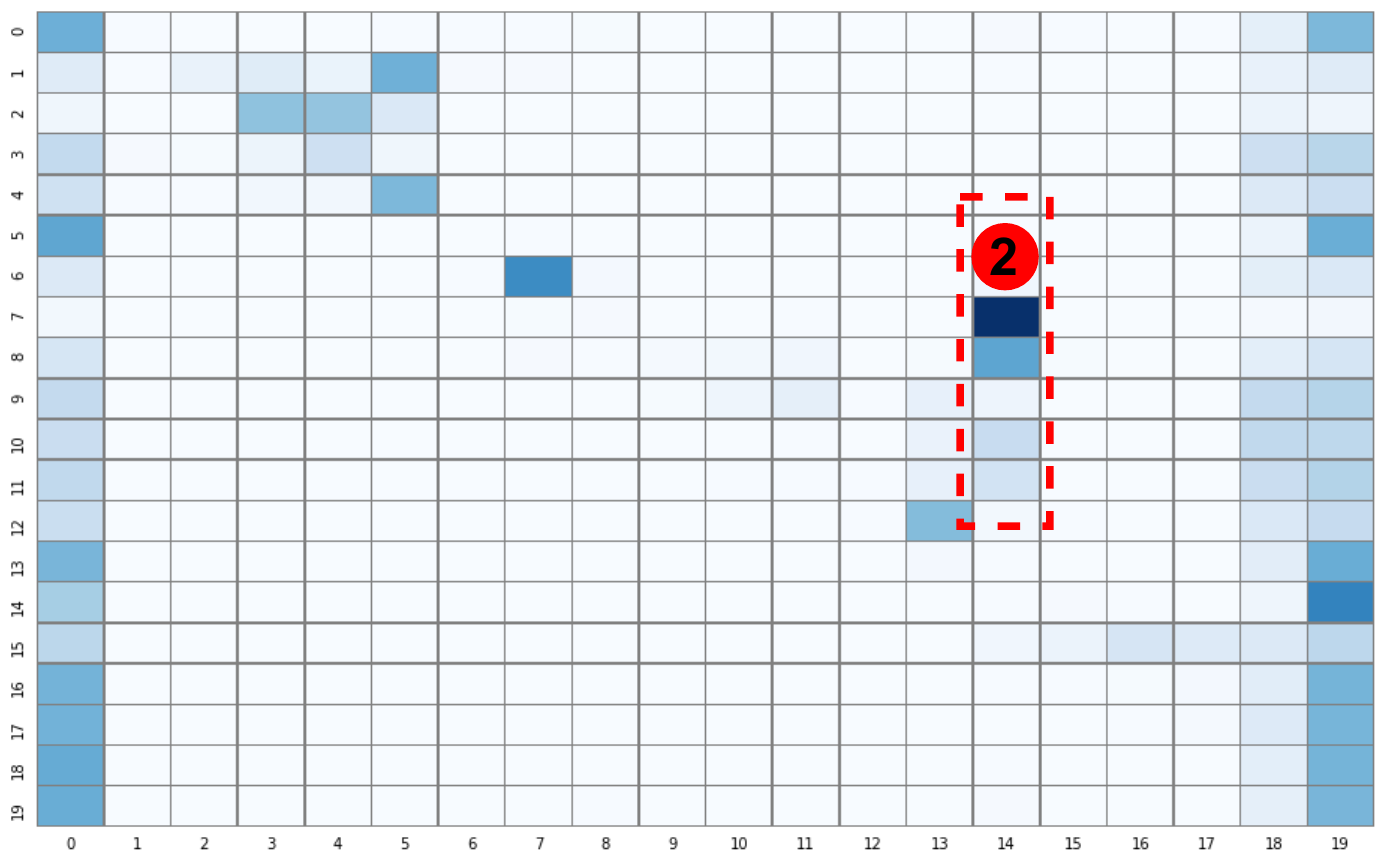}
            \caption[Middle Layer]%
            {{\small Middle Layer}}    
            \label{fig:attn3}
        \end{subfigure}
        \hfill
        \begin{subfigure}[b]{0.475\textwidth}   
            \centering 
            \includegraphics[width=\textwidth]{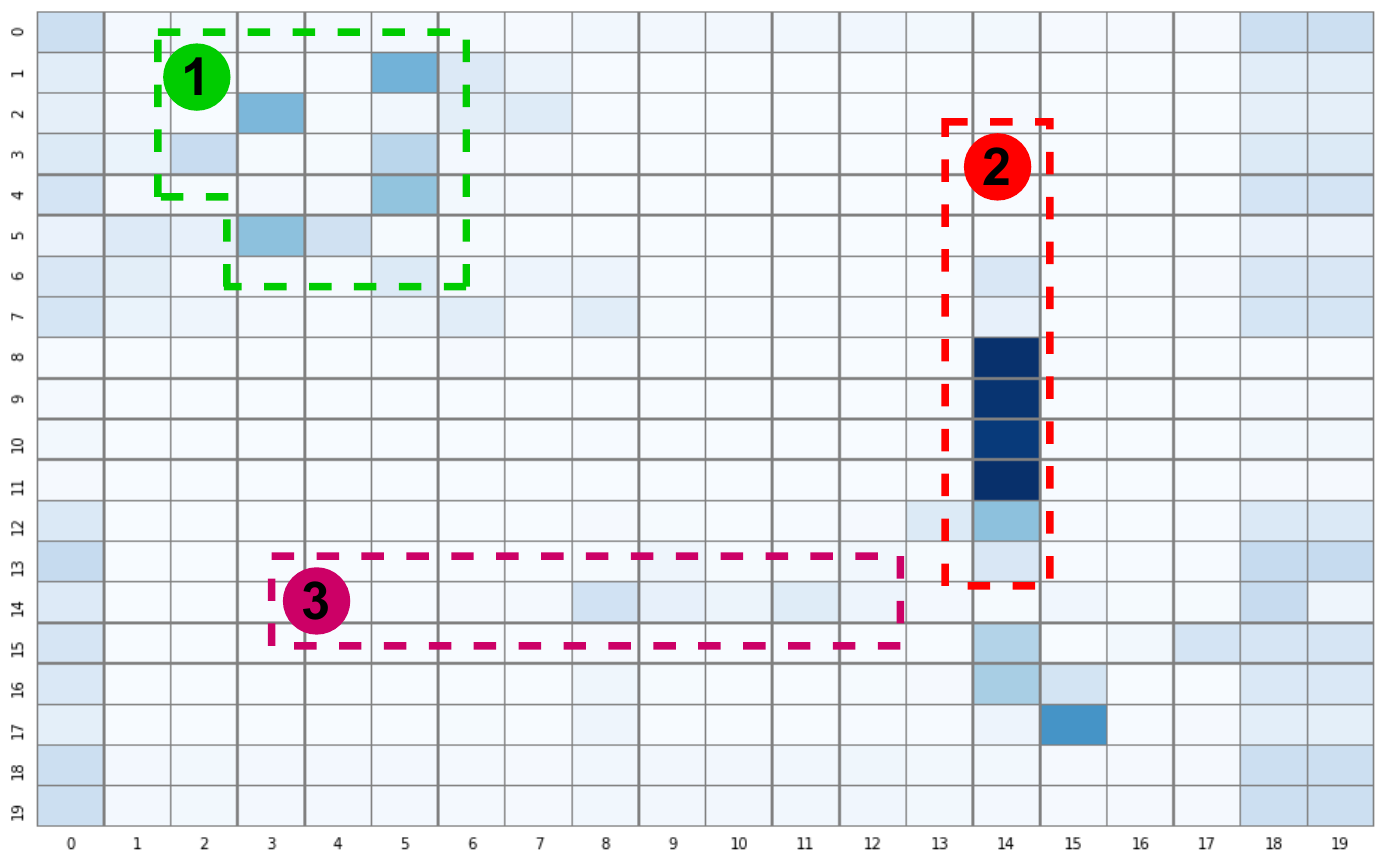}
            \caption[Last Layer]%
            {{\small Last Layer}}    
            \label{fig:attn4}
        \end{subfigure}
        \vspace{0.1cm}
        \end{minipage}
        \caption{The attention weights for the CoNLL 2009 example "[{\tt CLS}] The most troublesome report may be the August merchandise trade deficit due out tomorrow . [{\tt SEP}]". The first figure shows the dependency graph matrix.\label{fig:attn}} 
        \vspace{0.1cm}
    \end{figure*}

\begin{figure*}[htb!]
        \centering
        \begin{minipage}{0.6\linewidth}
        \begin{subfigure}[b]{0.475\textwidth}
            \centering
            \includegraphics[width=\textwidth]{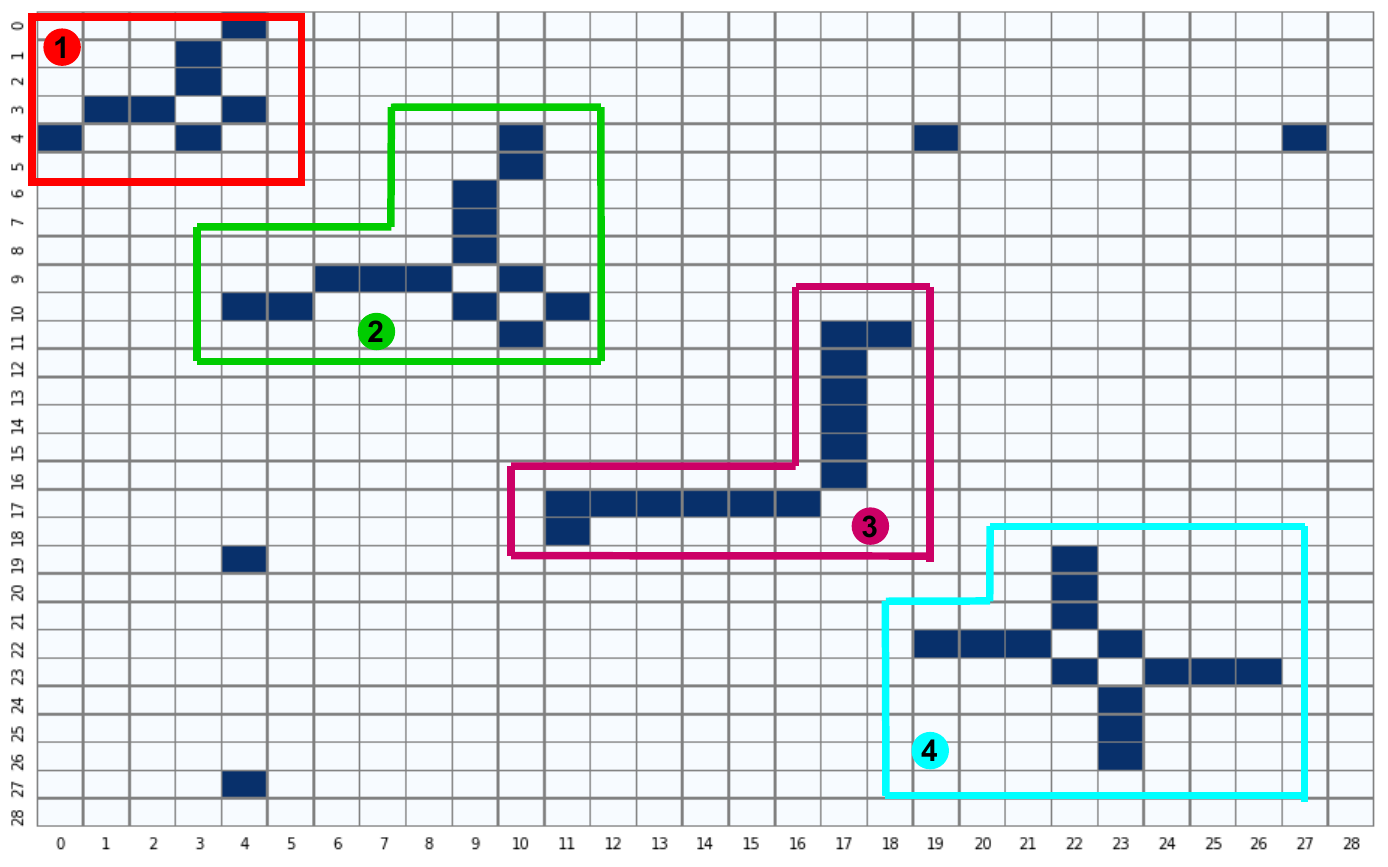}
            \caption[Graph Relation Matrix]%
            {{\small Graph Relation Matrix}}    
        \end{subfigure}
        \hfill
        \begin{subfigure}[b]{0.475\textwidth}  
            \centering 
            \includegraphics[width=\textwidth]{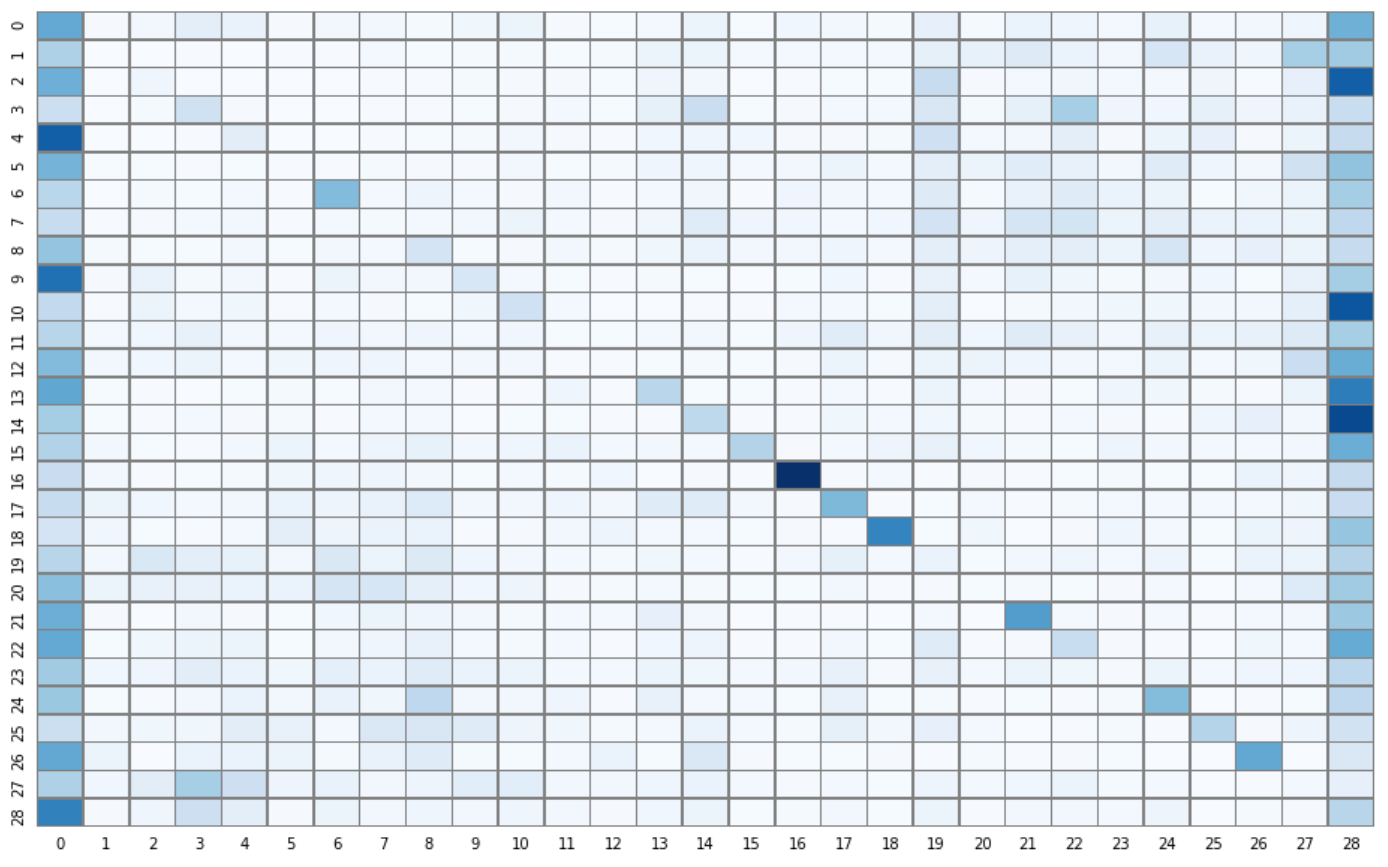}
            \caption[First Layer]%
            {{\small First Layer}}    
        \end{subfigure}
        \vspace{0.1cm}
        \begin{subfigure}[b]{0.475\textwidth}   
            \centering 
            \includegraphics[width=\textwidth]{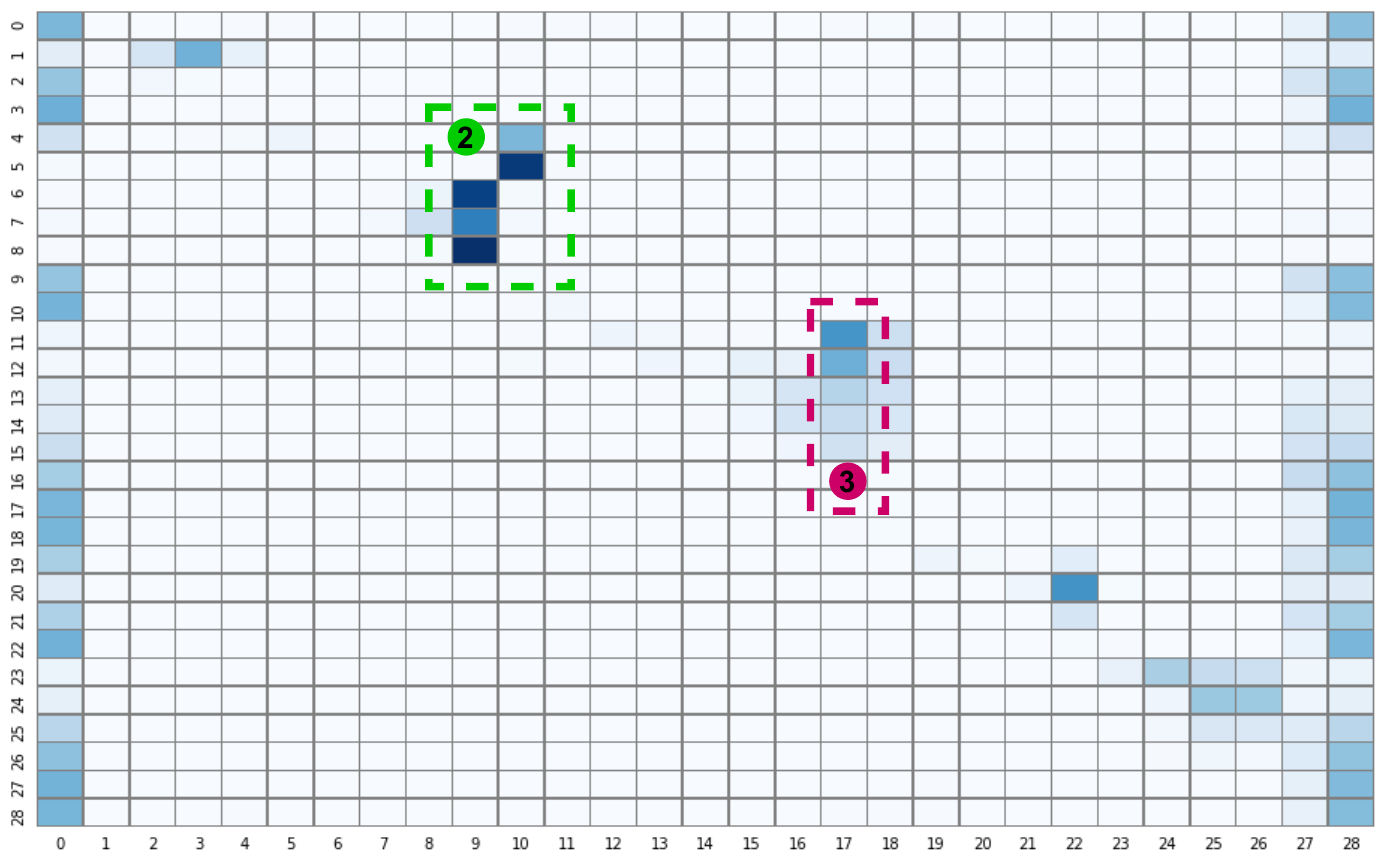}
            \caption[Middle Layer]%
            {{\small Middle Layer}}    
        \end{subfigure}
        \hfill
        \begin{subfigure}[b]{0.475\textwidth}   
            \centering 
            \includegraphics[width=\textwidth]{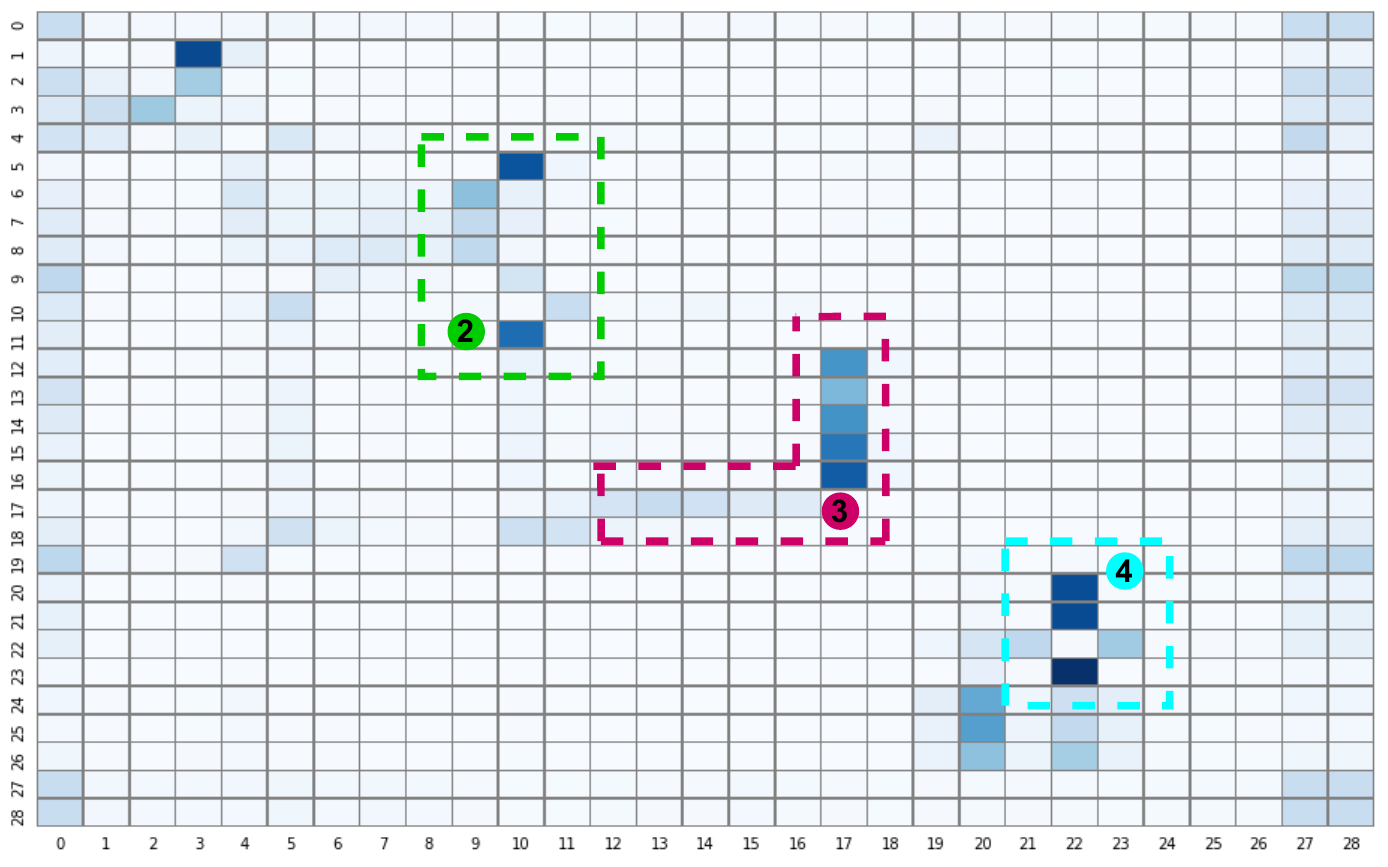}
            \caption[Last Layer]%
            {{\small Last Layer}}    
        \end{subfigure}
        \vspace{0.1cm}
        \end{minipage}
        \caption{The attention weights for a specific example "[{\tt CLS}] The consensus view expects a 0.4 \% increase in the September CPI after a flat reading in August . [{\tt SEP}]" on CoNLL 2009 dataset. The first figure shows the dependency graph matrix.\label{fig:attnt2}} 
        \vspace{0.1cm}
    \end{figure*}

\begin{figure*}[htb!]
        \centering
        \begin{minipage}{0.5\linewidth}
        \begin{subfigure}[b]{0.475\textwidth}
            \centering
            \includegraphics[width=\textwidth]{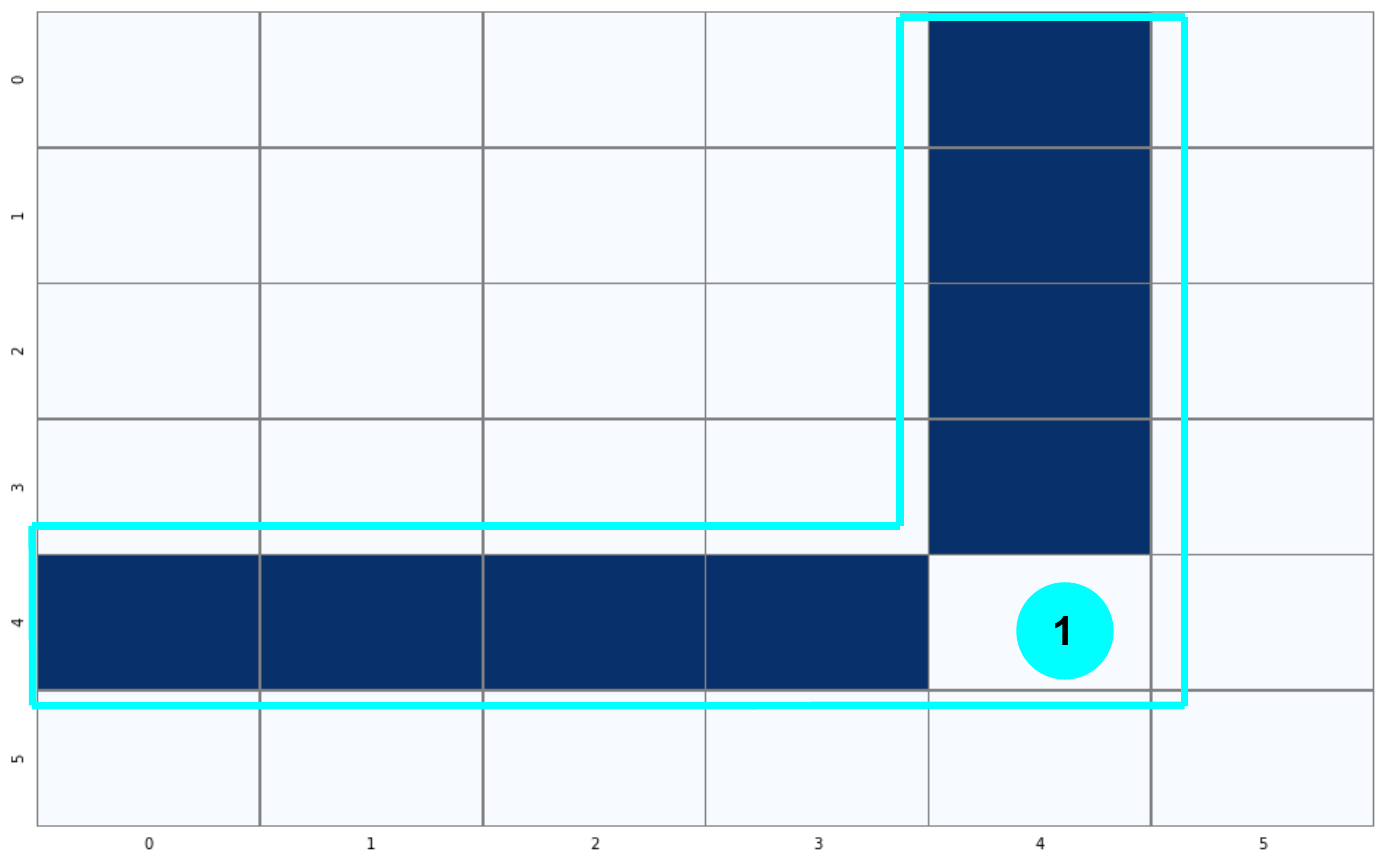}
            \caption[Graph Relation Matrix]%
            {{\small Graph Relation Matrix}}    
        \end{subfigure}
        \hfill
        \begin{subfigure}[b]{0.475\textwidth}  
            \centering 
            \includegraphics[width=\textwidth]{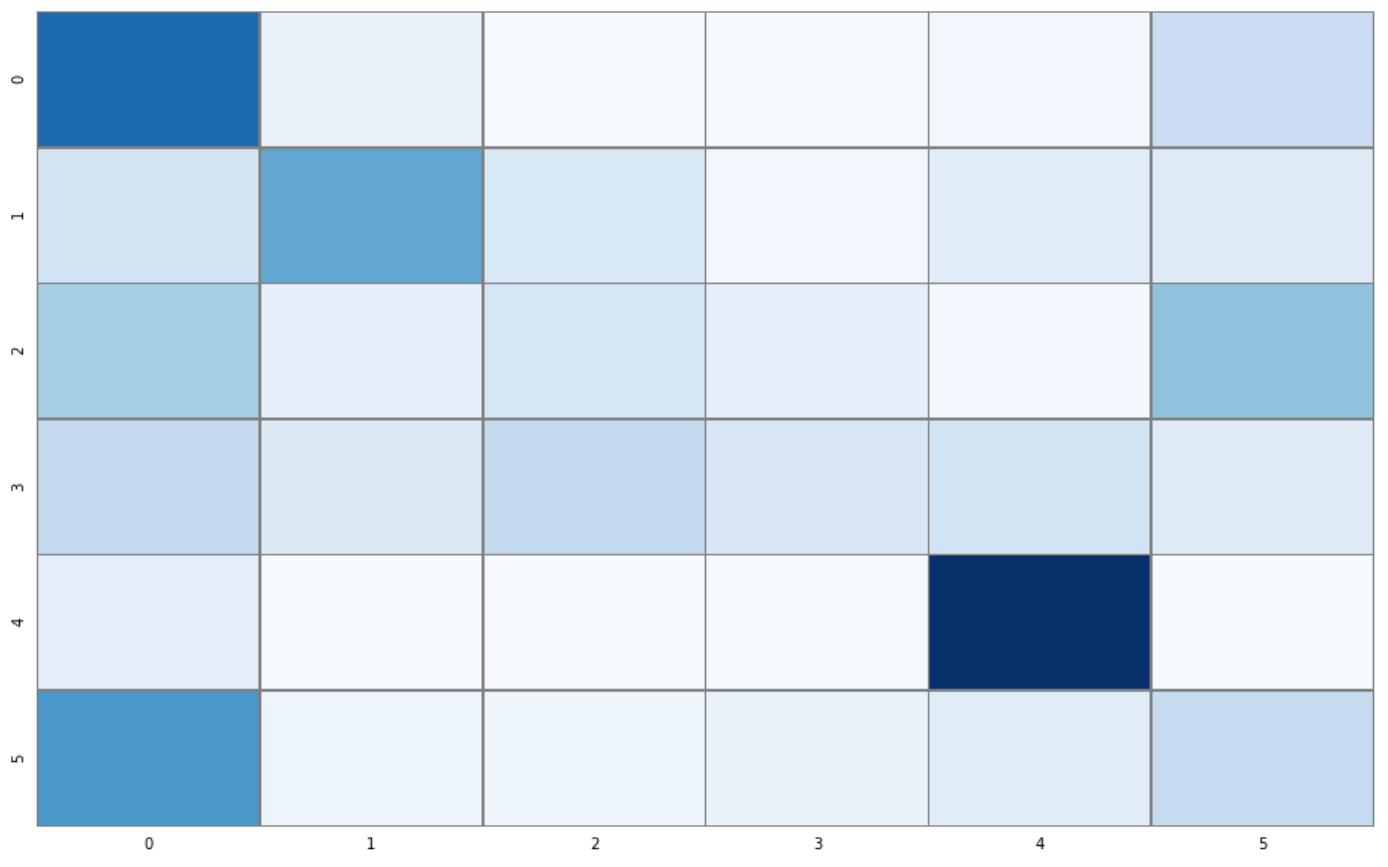}
            \caption[First Layer]%
            {{\small First Layer}}    
        \end{subfigure}
        \vspace{0.1cm}
        \begin{subfigure}[b]{0.475\textwidth}   
            \centering 
            \includegraphics[width=\textwidth]{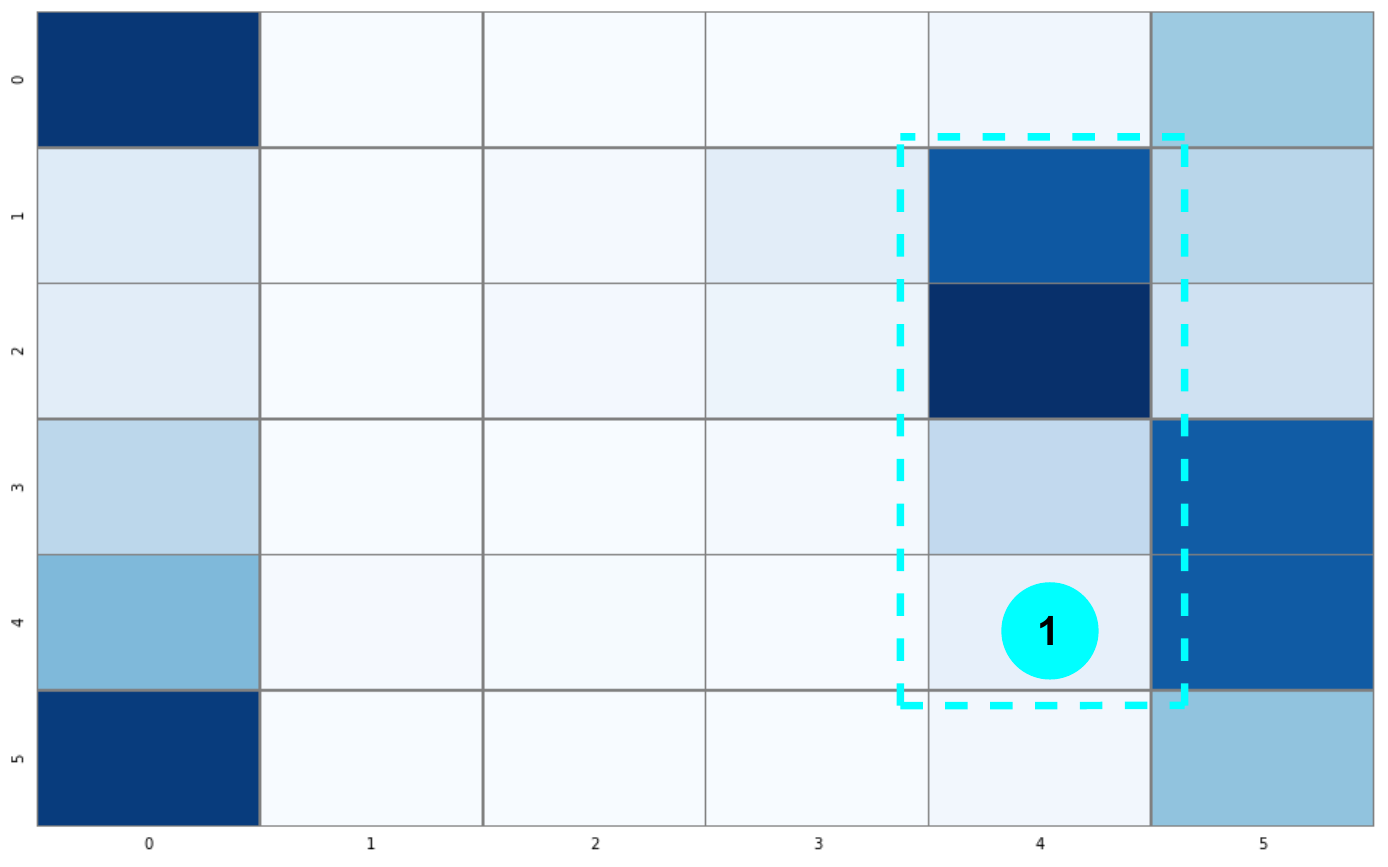}
            \caption[Middle Layer]%
            {{\small Middle Layer}}    
        \end{subfigure}
        \hfill
        \begin{subfigure}[b]{0.475\textwidth}   
            \centering 
            \includegraphics[width=\textwidth]{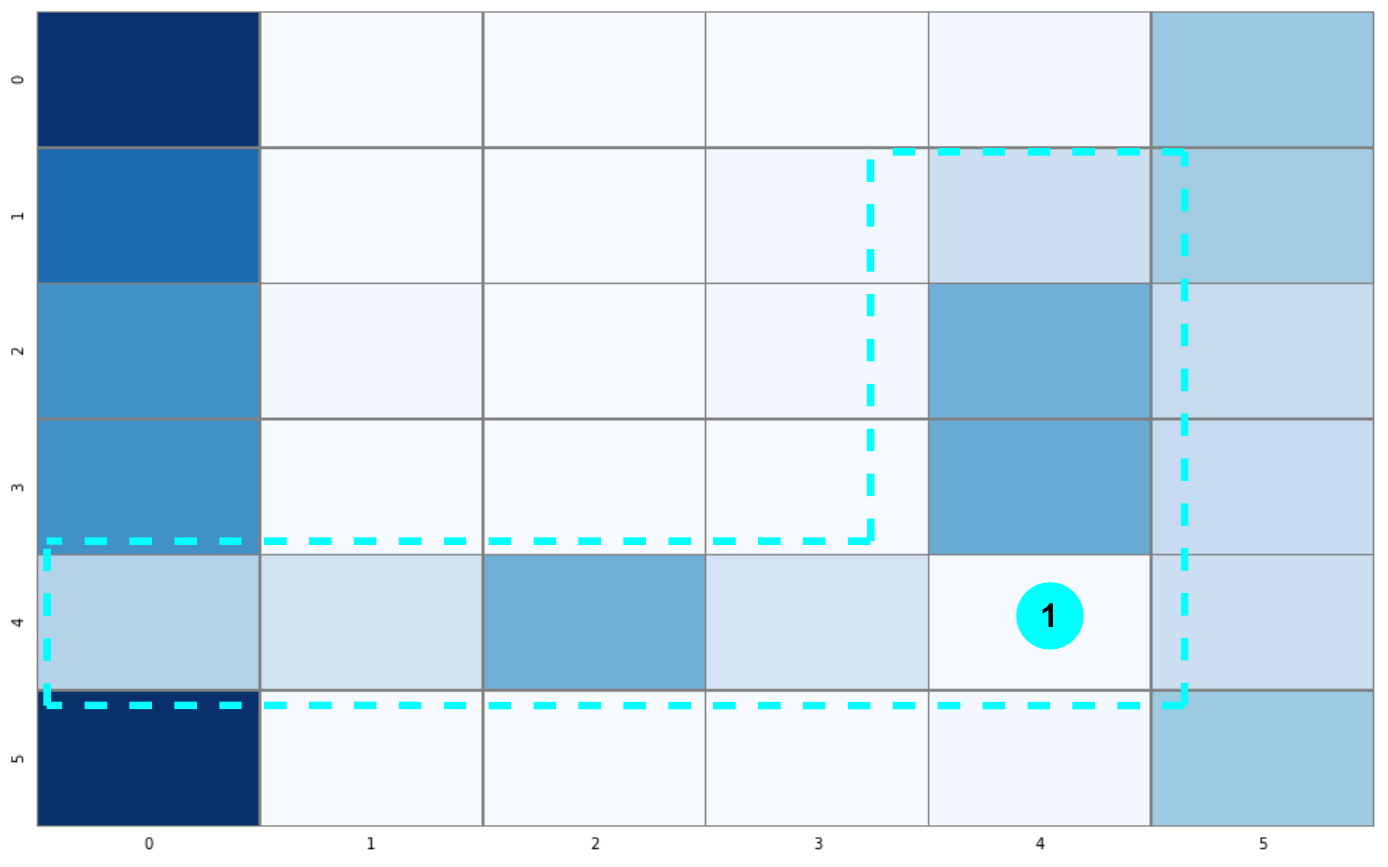}
            \caption[Last Layer]%
            {{\small Last Layer}}    
        \end{subfigure}
        \vspace{0.1cm}
        \end{minipage}
        \caption{The attention weights for a specific example "[{\tt CLS}] Candid Comment [{\tt SEP}]" on CoNLL 2009 dataset. The first figure shows the dependency graph matrix.\label{fig:attnt3}} 
        \vspace{0.1cm}
    \end{figure*}

\begin{figure*}[htb!]
        \centering
        \begin{minipage}{0.6\linewidth}
        \begin{subfigure}[b]{0.475\textwidth}
            \centering
            \includegraphics[width=\textwidth]{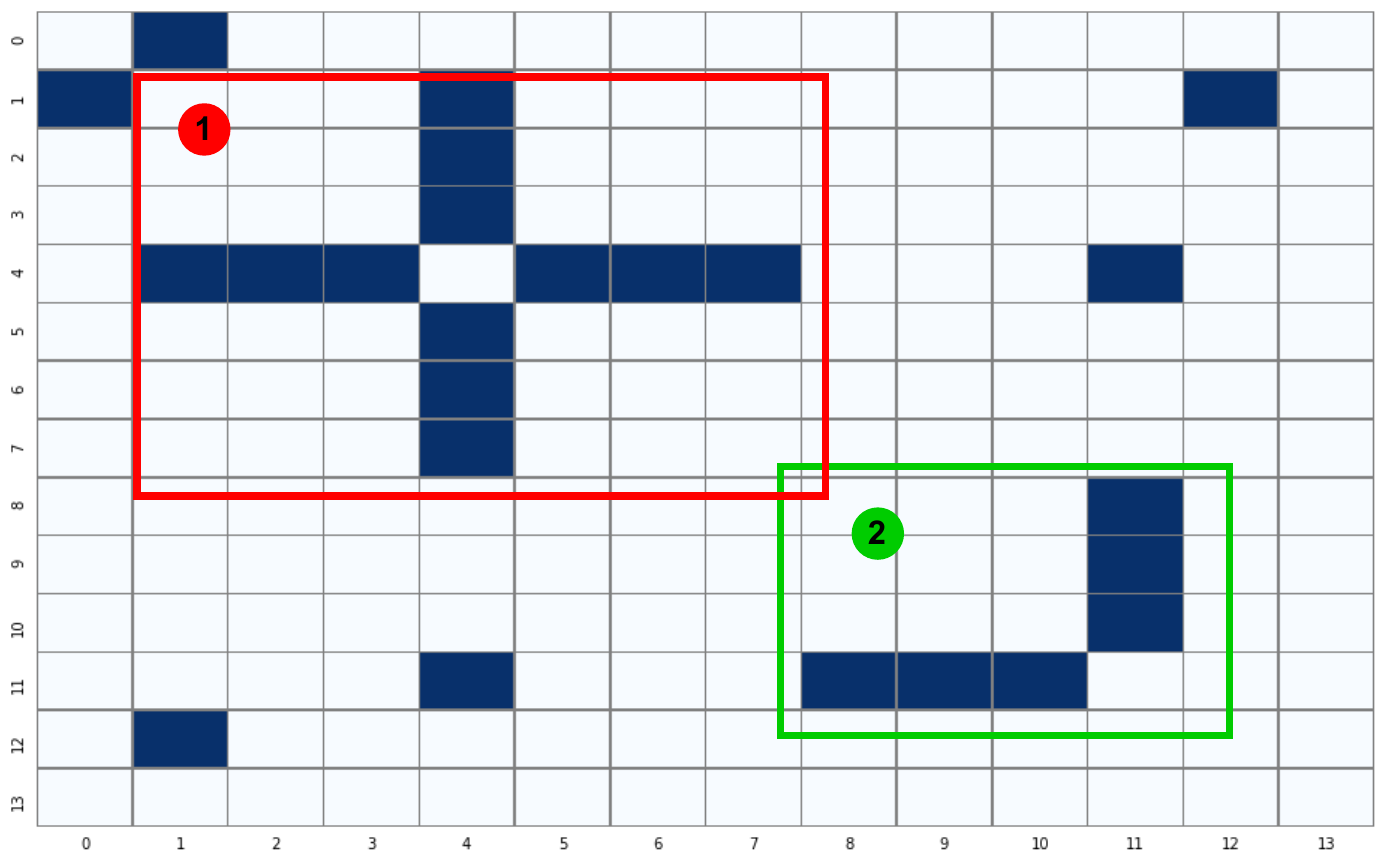}
            \caption[Graph Relation Matrix]%
            {{\small Graph Relation Matrix}}    
        \end{subfigure}
        \hfill
        \begin{subfigure}[b]{0.475\textwidth}  
            \centering 
            \includegraphics[width=\textwidth]{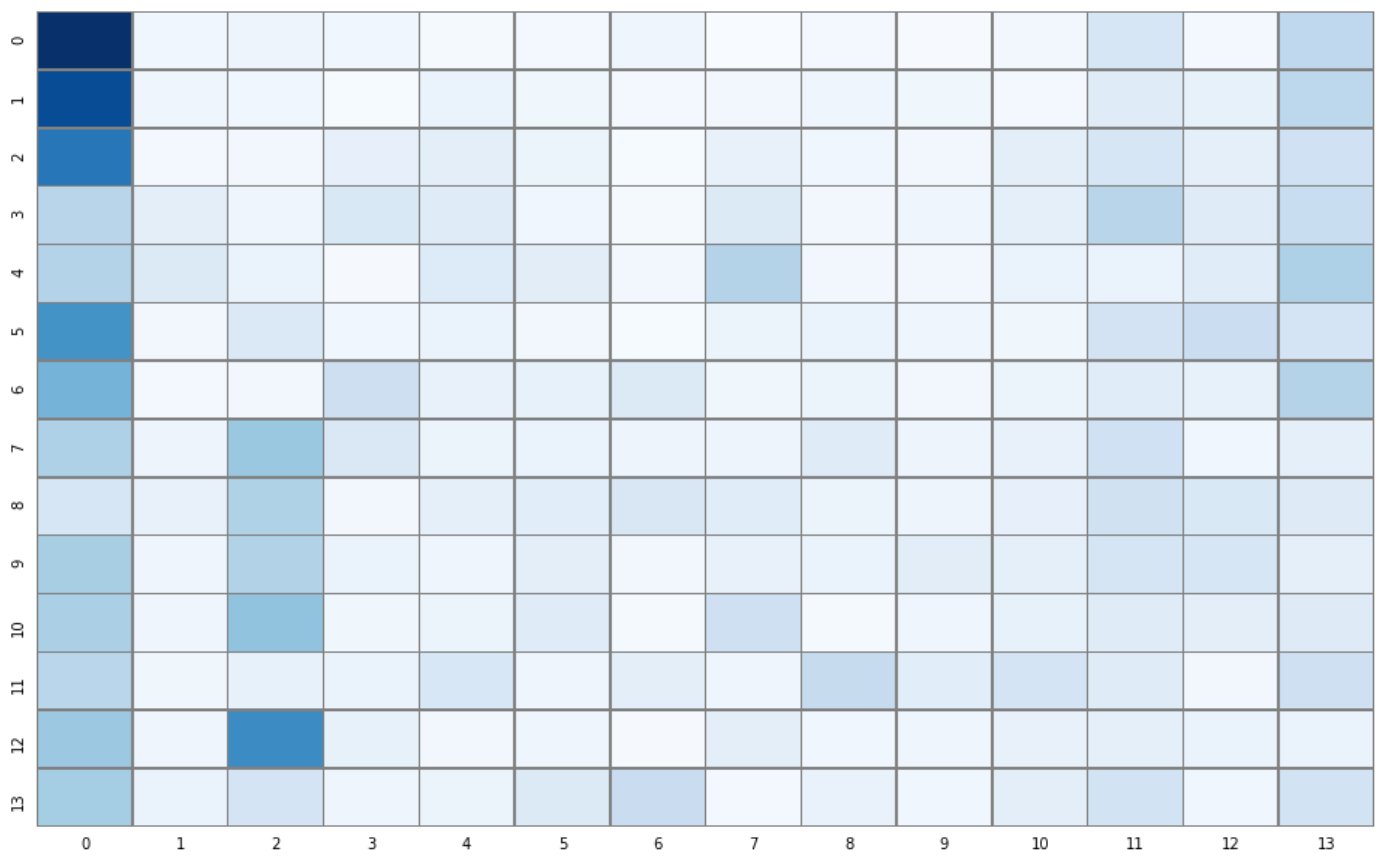}
            \caption[First Layer]%
            {{\small First Layer}}    
        \end{subfigure}
        \vspace{0.1cm}
        \begin{subfigure}[b]{0.475\textwidth}   
            \centering 
            \includegraphics[width=\textwidth]{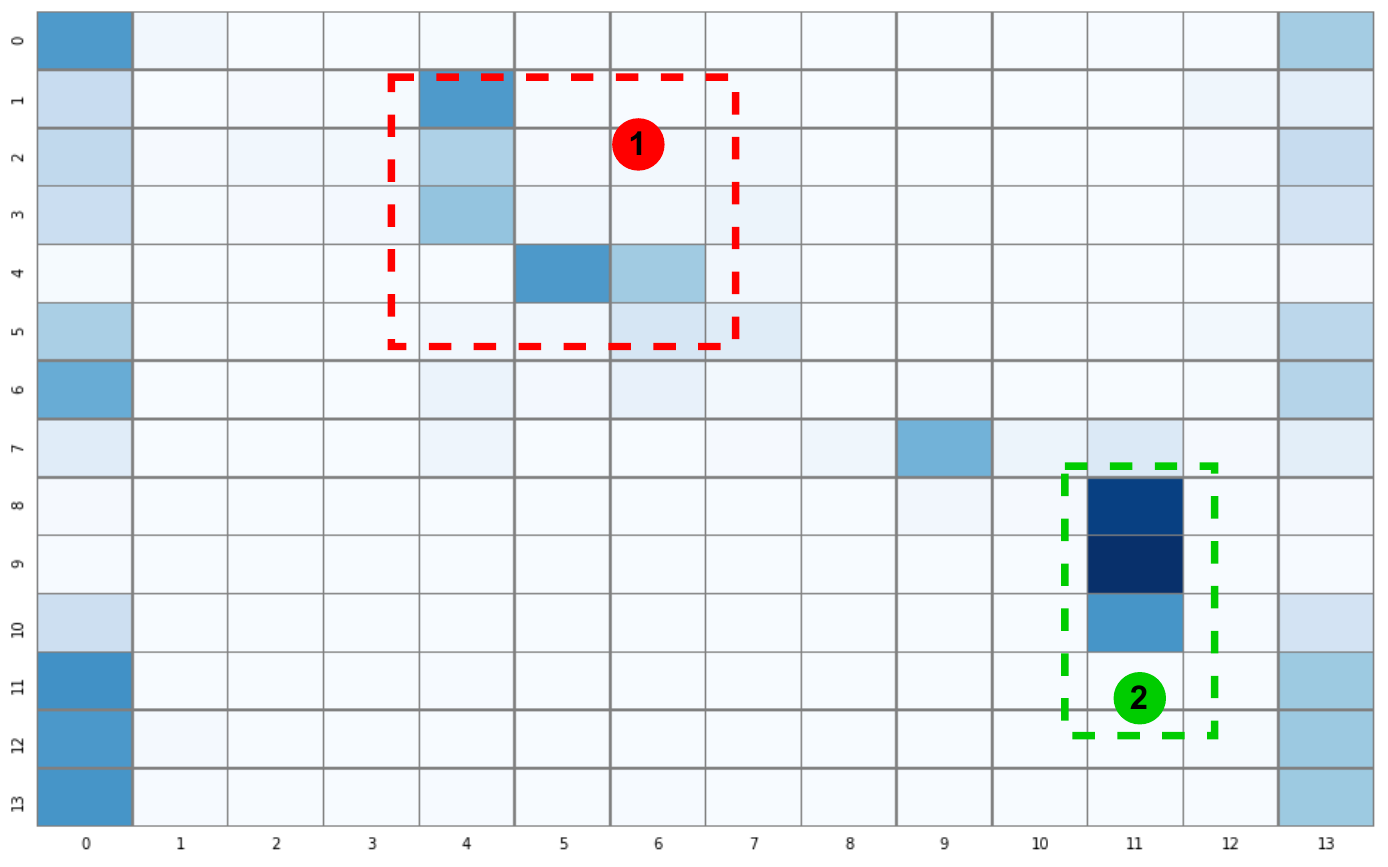}
            \caption[Middle Layer]%
            {{\small Middle Layer}}    
        \end{subfigure}
        \hfill
        \begin{subfigure}[b]{0.475\textwidth}   
            \centering 
            \includegraphics[width=\textwidth]{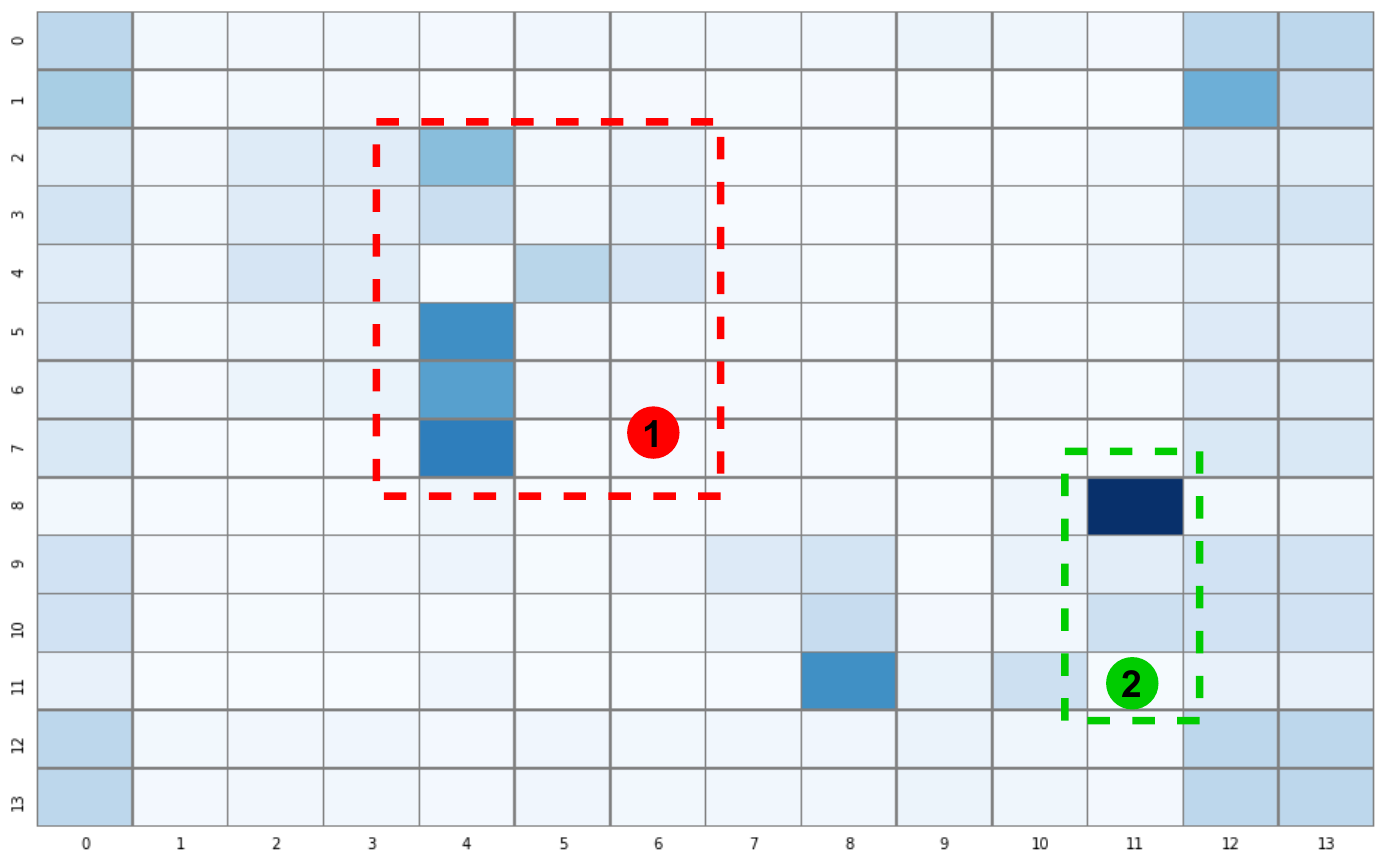}
            \caption[Last Layer]%
            {{\small Last Layer}}    
        \end{subfigure}
        \vspace{0.1cm}
        \end{minipage}
        \caption{The attention weights for a specific example "[{\tt CLS}] Let 's make that 1929 , just to be sure . [{\tt SEP}]" on CoNLL 2009 dataset. The first figure shows the dependency graph matrix.\label{fig:attnt4}} 
        \vspace{0.1cm}
    \end{figure*}

Figure~\ref{fig:attn} shows the attention weights for different layers of self-attention in the SynG2G-Tr model~(Figure \ref{fig:attn2}-\ref{fig:attn4}), alongside the dependency relation matrix~(Figure~\ref{fig:attn1}). The self-attention matrix includes four patterns. The first layer of the SynG2G-Tr model~(Figure~\ref{fig:attn2}) ignores the graph relations and learns string-local context information. For the middle layer~(Figure~\ref{fig:attn3}), attention weights partially use the graph relation pattern. Then, in the last layer~(Figure~\ref{fig:attn4}), the dependency graph relations are evident in the attention pattern. This demonstrates the benefit of adding the graph information with a soft bias, allowing the model to learn different structures using both local context and graph information. Furthermore, it can be inferred that the last layers of the self-attention mechanism require a global view and between-edge information, while the first few layers learn local context information. More examples are provided in Figures~\ref{fig:attnt2}-\ref{fig:attnt3}-\ref{fig:attnt4}. 

    \clearpage
    ~
    
\newpage

\section{Ablation Study}
\label{app:ablation}

In Table~\ref{srl-dev}, we analyse the interaction of the dependency graph with key and query vectors in the attention mechanism, as defined in Equation~\ref{eq:g2g-attn}. Excluding the key interaction results in a similar attention mechanism as defined in \newcite{mohammadshahi2020recursive}. This SynG2G-Tr\textit{-key} model achieves similar results compared to the SynG2G-Tr model on the WSJ test dataset given the predicate, but the SynG2G-Tr model outperforms it on all other settings, including both types of out-of-domain datasets, confirming that key interaction is a critical part of the SynG2G-Tr model. \\
When both key and query interactions are excluded from the SynG2G-Tr model~(SynG2G-Tr\textit{-key-query}), it has significantly lower performance than the SynG2G-Tr model in all settings.  This demonstrates the impact of encoding the graph relation embeddings in the self-attention mechanism of Transformer~\cite{transformervaswani} model. \\
We also evaluate adding the interaction of graph relations with value vectors to the SynG2G-Tr model, as defined in \newcite{espinosa-anke-etal-2022-multilingual,mohammadshahi-henderson-2020-graph}. The SynG2G-Tr\textit{+value} model achieves similar or worse results compared to the SynG2G-Tr model. So, we exclude this interaction to speed up the modified attention mechanism.

\begin{table}[htb!]
\centering
  \begin{adjustbox}{width=0.7\linewidth}
  \begin{tabular}{lcccccccc}
    \toprule
    \multirow{2}{*}{Model} &
      \multicolumn{3}{c}{CoNLL 2005} &&
      \multicolumn{3}{c}{CoNLL 2009} \\
      \cline{2-4} \cline{6-8}
    & Dev & WSJ & Brown && Dev & WSJ & Brown \\
    \midrule
    \textit{end-to-end}\\
    SynG2G-Tr \textit{-key} \textit{-query} & 86.65 & 87.08 & 79.40 && 86.40 & 87.26 & 81.12 \\
    SynG2G-Tr \textit{-key} & 86.82 & 87.27 & 80.33 && 86.85 & 87.50 & 81.51 \\
    SynG2G-Tr & 87.08 & 87.57 & 80.53 && 87.13 & 88.05 & 81.93 \\
    SynG2G-Tr \textit{+value} & 87.17 & 87.45 & 80.40 && 86.92 & 87.95 & 82.03 \\
    
    \hline
    \hline
    \textit{given predicate} \\
    SynG2G-Tr \textit{-key} \textit{-query} & 87.93 & 88.52 & 82.56 && 90.16 & 90.68 & 85.72 \\
    SynG2G-Tr \textit{-key} & 88.03 & 88.91 & 82.90 && 90.31 & 91.22 & 86.28 \\
    SynG2G-Tr &  88.17 & 88.93 & 83.21 && 90.66 & 91.23 & 86.43 \\
    SynG2G-Tr \textit{+value} & 88.15 & 88.78 & 83.10 && 90.48 & 91.15 & 86.41 \\
    \bottomrule
  \end{tabular}
  \end{adjustbox}
  \caption{Model comparison of SynG2G-Tr and other variants, by F1 score on CoNLL 2005 and CoNLL 2009 datasets. The SynG2G-Tr\textit{-key-query} model is the same as the syntax-agnostic BERT model.\label{srl-dev}
  }
\end{table}

\end{appendices}

\end{document}